\documentclass[conference]{IEEEtran}

\usepackage{xspace}
\usepackage{microtype}
\usepackage{graphicx}
\usepackage{subfigure}
\usepackage{amsmath}
\usepackage{booktabs} 
\usepackage{graphicx,subfigure}
\usepackage{makecell}
\usepackage{enumitem}
\usepackage{url}
\usepackage{amsfonts}
\usepackage[linesnumbered,ruled,vlined]{algorithm2e}
\usepackage[hidelinks]{hyperref}

\def\BibTeX{{\rm B\kern-.05em{\sc i\kern-.025em b}\kern-.08em
    T\kern-.1667em\lower.7ex\hbox{E}\kern-.125emX}}

\newcommand*{\eg}{e.g.,\xspace}
\newcommand*{\ie}{i.e.,\xspace}

\newcommand*{\name}{\texttt{P4}\xspace}
\newcommand*{\cifar}{\texttt{CIFAR-10}\xspace}
\newcommand*{\cifarext}{\texttt{CIFAR-100}\xspace}
\newcommand*{\femnist}{\texttt{FEMNIST}\xspace}
\newcommand*{\lf}{\textit{label flipping}\xspace}
\newcommand*{\byz}{\textit{byzantine-zero}\xspace}
\newcommand*{\byzr}{\textit{byzantine-random}\xspace}
\newcommand*{\byzf}{\textit{byzantine-flip}\xspace}
\newcommand*{\mk}{\texttt{m-Krum}\xspace}
\newcommand*{\anom}{\texttt{anomaly-detection}\xspace}

\newcommand{\submit}[1]{}
\newcommand{\arxiv}[1]{#1}

\begin{document}

\title{Client Clustering Meets Knowledge Sharing: Enhancing Privacy and Robustness in Personalized Peer-to-Peer Learning}

\author{\IEEEauthorblockN{Mohammad M Maheri, Denys Herasymuk, Hamed Haddadi}
\IEEEauthorblockA{
Imperial College London}
}

\maketitle

\begin{abstract}
The growing adoption of Artificial Intelligence (AI) in Internet of Things (IoT) ecosystems has intensified the need for personalized learning methods that can operate efficiently and privately across heterogeneous, resource-constrained devices. However, enabling effective personalized learning in decentralized settings introduces several challenges, including efficient knowledge transfer between clients, protection of data privacy, and resilience against poisoning attacks. In this paper, we address these challenges by developing \name (Personalized, Private, Peer-to-Peer)---a method designed to deliver personalized models for resource-constrained IoT devices while ensuring differential privacy and robustness against poisoning attacks. Our solution employs a lightweight, fully decentralized algorithm to privately detect client similarity and form collaborative groups. Within each group, clients leverage differentially private knowledge distillation to co-train their models, maintaining high accuracy while ensuring robustness to the presence of malicious clients. We evaluate \name on popular benchmark datasets using both linear and CNN-based architectures across various heterogeneity settings and attack scenarios.  Experimental results show that \name achieves 5\% to 30\% higher accuracy than leading differentially private peer-to-peer approaches and maintains robustness with up to 30\% malicious clients. Additionally, we demonstrate its practicality by deploying it on resource-constrained devices, where collaborative training between two clients adds only $\approx 7$ seconds of overhead.

\end{abstract}


\section{Introduction}
\label{sec:introduction}

The integration of Artificial Intelligence (AI) into Internet of Things (IoT) ecosystems is reshaping how connected devices operate by enabling autonomous decision-making and real-time data analysis at scale. In healthcare, for example, AI empowers clinical staff to continuously monitor patients’ physiological data throughout the hospital, detect early signs of deterioration, and issue timely alerts---ultimately helping to prevent adverse outcomes. The diverse nature of data---for example, significant differences in electrocardiograms resulting from varying levels of patient health---collected by different devices is known as \textit{client data heterogeneity}. This heterogeneity often causes a single global model to perform poorly across varying data distributions since each client optimizes towards its own distinct local empirical minimum, leading to divergent local optimization directions\submit{~\cite{li2022towards}}\arxiv{~\cite{li2022towards,qu2022generalized,li2021model,luo2021no}}. Recently, growing interest has emerged in addressing data heterogeneity using \textit{personalized learning}~\submit{\cite{collins2021exploiting}}\arxiv{\cite{li2022learning,collins2021exploiting,li2021ditto}}, which adapts each client’s model to its specific data or task while still utilizing shared knowledge across clients.

Fully decentralized or \textit{peer-to-peer} (P2P) learning has emerged as a promising direction for enabling personalized learning in heterogeneous environments such as those found in IoT ecosystems. In contrast to centralized federated learning (FL), peer-to-peer learning eliminates the need for a central server and distributes computation among participating clients~\submit{\cite{li2022learning}}\arxiv{\cite{zantedeschi2020fully, li2022learning}}. This approach leverages direct communication between clients with their own data distributions, offering potential benefits such as faster convergence\submit{~\cite{li2022learning}}\arxiv{~\cite{sun2022decentralized, li2022learning}} and enhanced data privacy compared to centralized federated learning~\submit\cite{shi2023improving}\arxiv{\cite{shi2023improving, kairouz2021advances}}. In the context of IoT, where devices often operate under bandwidth and power constraints, P2P learning allows edge devices to collaboratively train models by communicating only with a limited number of neighbors, depending on their communication and computational capabilities\submit{~\cite{lalitha2019peer}}\arxiv{~\cite{lalitha2019peer,warnat2021swarm}}.

\textit{Personalized decentralized learning} faces three key challenges. The first is \textit{clustering}, where clients seek similar peers to share knowledge. Clustering algorithms~\submit{\cite{li2022towards}}\arxiv{\cite{li2022towards,duan2021flexible}} group clients with similar data distributions to train localized models, improving accuracy. However, most methods rely on predefined cluster counts and require a central server\arxiv{~\cite{xie2021multi,sattler2020clustered,nguyen2022self,briggs2020federated}}, which is impractical due to communication bandwidth constraints. More critically, these approaches risk privacy leaks as clients share data for clustering. The second challenge is \textit{data privacy}---clients' data often contains sensitive information, requiring strong privacy guarantees. A common solution is adding differential privacy (DP) noise and clipping gradients\submit{~\cite{li2019asynchronous}}\arxiv{~\cite{li2019asynchronous, wei2020federated, truex2020ldp, geyer2017differentially}}, but this degrades model performance\submit{~\cite{bagdasaryan2019differential}}\arxiv{~\cite{bagdasaryan2019differential,noble2022differentially,wei2020federated}}. In the peer-to-peer settings, this effect is amplified due to fewer model updates (gradients) that are used in model aggregation~\cite{kalra2023decentralized}. Finally, the third challenge is the presence of \textit{malicious clients} affected by data poisoning\arxiv{~\cite{tolpegin2020data}} or model poisoning\arxiv{~\cite{lin2019free, xu2022byzantine, fang2020local, chen2017distributed}} attacks\submit{~\cite{han2024fedsecurity}}, which can degrade the final model performance. Due to its fully decentralized nature, peer-to-peer learning is highly vulnerable to poisoning attacks, in which malicious clients can exploit the system by sending carefully crafted local models to their neighboring peers. A common mitigation strategy is to use recent decentralized FL (DFL) defenses~\submit{\cite{fang2024byzantine}}\arxiv{\cite{fang2024byzantine, heydaribeni2023surefed, sun2024byzantine}} or adapt existing centralized FL ones~\submit{\cite{han2024fedsecurity}}\arxiv{\cite{sun2019can, yin2018byzantine, blanchard2017machine, fung2020limitations}} to DFL. However, a defense that is not tailored to personalized peer-to-peer settings may fail in scenarios with highly heterogeneous clients and can introduce additional computational overhead, particularly when each client is required to perform robust aggregation individually~\cite{hallaji2024decentralized}.

\arxiv{Although recent efforts in the FL community have introduced privacy into the P2P setting~\submit{\cite{kalra2023decentralized, bayrooti2023differentially}}\arxiv{\cite{kalra2023decentralized, bayrooti2023differentially, shayan2020biscotti, arapakis2023p4l}}, to the best of our knowledge, no existing fully decentralized framework explicitly addresses the challenge of ensuring differential privacy under heterogeneous client data distributions typical of IoT environments. There is a growing trend of applying differential privacy to improve the privacy-utility trade-off in the P2P setting~\cite{kalra2023decentralized, bayrooti2023differentially}. However, these methods are less effective when clients have highly heterogeneous data (see Section~\ref{sec:privacy_utility}) and typically assume that clients are only curious about other clients' data while honestly sharing their local model gradients. While recent frameworks~\submit{\cite{mondal2022beas, qin2024blockdfl, shayan2020biscotti, arapakis2023p4l}}\arxiv{\cite{mondal2022beas, qin2024blockdfl, shayan2020biscotti, arapakis2023p4l, chen2018machine}} make progress in integrating privacy preservation and resilience to data poisoning in P2P settings, they suffer from several limitations. Most notably, they neglect client heterogeneity, offer limited improvements in the privacy-utility trade-off, and often fail to address model poisoning attacks, which are particularly threatening in DFL~\cite{fang2024byzantine}. However, in scenarios with highly data-heterogeneous clients where personalization is important, incorporating the similarity of the client's data distribution into the training procedure can improve both privacy-utility amplification~\submit{\cite{sattler2020clustered}}\arxiv{\cite{ghosh2020efficient,sattler2020clustered}} and robustness against poisoning attacks~\cite{han2023kick}.}

\submit{\vspace{\baselineskip}}

In this paper, we propose \name (Personalized, Private, Peer-to-Peer)\footnote{\url{https://github.com/mammadmaheri7/private_decentralized_learning}}, a method designed to deliver personalized models for resource-constrained IoT devices while ensuring differential privacy and robustness against poisoning attacks. In \name, clients first group themselves in a fully decentralized manner using a similarity metric based on their model weights. Once grouped, clients perform collaborative training with upper-bounded data privacy risk and robust aggregation against adversarial model updates. To enhance privacy-utility trade-offs, \name incorporates the proxy model~\cite{zhang2018deep} to facilitate knowledge distillation among clients. This separation between the local and proxy models results in fast convergence and personalized parameter adaptation for each client (as shown in Section~\ref{sec:privacy_utility}), while also enhancing resilience against data poisoning attacks (as discussed in Section~\ref{subsection:malicious_security}). To mitigate model poisoning attacks, \name employs a combined defense strategy based on \mk~\cite{blanchard2017machine} and \texttt{anomaly-detection}~\cite{han2023kick}, which utilizes client similarity within each group to filter out malicious updates. \name is well-suited for IoT deployments that involve numerous clients and non-stationary data distributions, where communication and computational efficiency are critical\arxiv{~\cite{dhar2021survey}}.

\submit{\vspace{0.15cm}}
\arxiv{\vspace{1.0cm}}

\noindent Our main contributions are summarized as follows:

\begin{itemize}[leftmargin=*]
    \item We design \name to address both data heterogeneity and poisoning threats in a differential private P2P setting tailored for IoT environments. \name enables clients with similar data distributions to collaboratively train, leveraging KL divergence-based regularization between local and proxy models to facilitate knowledge transfer. Its defense mechanism uniquely incorporates knowledge distillation, anomaly detection, and \mk to improve robustness.

    \item We propose a lightweight and decentralized procedure that allows clients to privately group themselves for collaborative training, using the $\ell_1$-norm between their model weights as a similarity metric. 

    \item We conduct a comprehensive empirical study of personalized P2P learning under differential privacy and adversarial attacks, covering a range of privacy budgets, model architectures, data heterogeneity types, and poisoning scenarios. Our results show that \name consistently outperforms state-of-the-art baselines by 5\%–30\% in accuracy and tolerates up to 30\% malicious clients, demonstrating both strong privacy-utility trade-offs and robustness.

    \item To evaluate the feasibility of deploying \name in real-world IoT settings, we implement it on resource-constrained edge devices using Raspberry Pis. Our results show minimal overhead in runtime, memory usage, power consumption, and communication bandwidth, with collaborative training between two clients adding only $\approx$~7 seconds of~overhead.

\end{itemize}

\submit{\section{Related Work}
\label{sec:related_work}

Most existing peer-to-peer learning approaches rely on random or predefined communication topologies, without evaluating whether the selected links are actually beneficial for collaboration~\submit{\cite{dai2022dispfl, shi2023improving}}\arxiv{\cite{shi2023towards, dai2022dispfl, shi2023improving, kalra2023decentralized, wang2023enhancing}}. While \cite{jeong2023personalized} attempted to build a dynamic topology based on client similarity, their method only considers output logit similarity and does not account for the underlying data distribution. This highlights the need for more intelligent and adaptive strategies for link selection that can improve collaboration effectiveness and accelerate convergence in environments with heterogeneous client data.

While differential privacy has been widely explored in centralized FL~\cite{noble2022differentially}, a significant gap remains in addressing both data and model privacy in P2P learning settings~\cite{dai2022dispfl, jeong2023personalized, shi2023improving}. Several approaches have been proposed to enhance privacy in peer-to-peer learning. \cite{bellet2018personalized} investigated the trade-off between utility and privacy in peer-to-peer FL. \cite{kalra2023decentralized} introduced proxy models to enable efficient communication without a central server and incorporated differential privacy analysis to strengthen privacy guarantees. However, the added noise from differential privacy can degrade model performance, particularly when client data is non-IID, and merging models from divergent distributions may further harm accuracy. More recently, \cite{bayrooti2023differentially} extended DP-SGD to decentralized learning, aiming to reach consensus on model parameters across clients. Yet, this consensus model may underperform in non-IID scenarios where personalized models are more appropriate. Another direction, explored by \cite{wang2023enhancing}, applied gradient encryption to protect privacy and used succinct proofs to verify gradient correctness. Despite its security benefits, this approach introduced computational and communication overhead and did not address data heterogeneity. These limitations underscore the need for future research on privacy-preserving mechanisms that address heterogeneity and personalization in P2P learning.

Several recent works\submit{~\cite{mondal2022beas, qin2024blockdfl, shayan2020biscotti, arapakis2023p4l}}\arxiv{\cite{mondal2022beas, qin2024blockdfl, shayan2020biscotti, arapakis2023p4l, chen2018machine}} have proposed frameworks for P2P learning that assume a stronger threat model, addressing both privacy and malicious security concerns. For instance, Biscotti~\cite{shayan2020biscotti} is a fully decentralized P2P system for privacy-preserving and secure federated learning, leveraging blockchain, differential privacy, and m-Krum~\cite{blanchard2017machine} for secure aggregation. However, this framework is limited to IID data and does not address client heterogeneity in the context of privacy preservation and robustness against poisoning attacks. These limitations reveal a significant gap in DFL frameworks: the lack of solutions that simultaneously support data-heterogeneous clients and operate under a strong threat model encompassing both privacy preservation and protection against data poisoning and model poisoning~attacks.
}\arxiv{\section{Background and Related Work}
\label{sec:related_work}

\subsection{Private Decentralized Learning}
\label{subsection:private_fl}

In decentralized learning, preserving both data and local model privacy is essential, as deep neural networks can unintentionally memorize training data, making them vulnerable to model inversion attacks~\cite{geiping2020inverting, huang2021evaluating}. To address this, several privacy-preserving techniques have been developed to protect plaintext gradients. Differential privacy has been applied to mitigate privacy risks~\cite{li2019asynchronous, wei2020federated, truex2020ldp, geyer2017differentially}, but the added noise can degrade model performance~\cite{truex2019hybrid}. Homomorphic encryption offers an alternative by encrypting model weights or gradients~\cite{aono2017privacy, zhang2020privacy, zhao2022pvd}, though it introduces significant computational and communication overhead~\cite{jin2023fedml}, limiting its practicality for resource-constrained edge devices. Secure multi-party computation (MPC) techniques~\cite{bonawitz2017practical, mandal2018nike, zheng2019helen} preserve privacy through collaborative computation among multiple parties, while keeping their individual inputs confidential, but are similarly hindered by high communication costs. More recently, trusted execution environments (TEEs) have been explored~\cite{mo2021ppfl, mo2022sok, dhasade2022tee}, though their reliance on specialized hardware and associated computational overhead restrict widespread adoption.

\subsection{Personalized Decentralized Learning}
\label{subsection:personalized_fl}

The presence of non-IID client data necessitates personalization in decentralized learning, shifting focus from a single global model to customized models for each client. \cite{li2021ditto} used regularization to keep personalized models close to the global optimum, while \cite{zhang2020personalized} proposed computing an optimal weighted combination of client models for personalized updates. Another line of work clusters clients to improve performance~\cite{ghosh2020efficient, sattler2020clustered}, with aggregation performed within clusters to avoid negative transfer that may result from merging dissimilar models~\cite{li2022towards}. A key challenge in federated clustering is measuring client similarity, typically based on losses~\cite{ghosh2020efficient, li2022towards}, gradients~\cite{sattler2020clustered, duan2021flexible}, or model weights~\cite{xie2021multi, sattler2020clustered, nguyen2022self, briggs2020federated}. However, aside from \cite{li2022towards}, these methods rely on a centralized server, limiting their applicability in P2P settings. Moreover, most approaches, including \cite{li2022towards}, overlook privacy risks during similarity computation. Loss-based methods, in particular, require each client to receive other clients' data and compute the loss of its data on the received model to compute its similarity with other clients, incurring high communication and computation overhead. Moreover, most approaches assume prior knowledge of the number of client clusters, leading to degraded clustering performance if this number is mis-estimated.

\subsection{P2P Learning}
\label{subsection:decentralized_fl}

Several studies have explored decentralized federated learning and proposed techniques to enhance its performance. Some employed Bayesian methods to model shared knowledge among clients in a decentralized setting~\cite{lalitha2018fully, lalitha2019peer}, while \cite{roy2019braintorrent} introduced a server-less federated learning framework designed for dynamic environments. To reduce communication and computation overhead, \cite{sun2022decentralized} applied quantization techniques, and \cite{dai2022dispfl} leveraged sparse training methods to achieve similar efficiency~gains.

To address the challenge of heterogeneous data distributions in P2P learning, several personalization strategies have been proposed. \cite{dai2022dispfl} used personalized sparse masks to tailor models to individual clients. \cite{shi2023towards} introduced a decentralized partial model training approach using the Sharpness Aware Minimization (SAM) optimizer~\cite{foret2020sharpness} to mitigate model inconsistency. This was further improved by combining SAM with Multiple Gossip Steps (MGS) to address both inconsistency and overfitting~\cite{shi2023improving}. \cite{jeong2023personalized} tackled data heterogeneity by employing weighted aggregation of model parameters based on the Wasserstein distance between output logits of neighboring clients.

However, most existing peer-to-peer learning approaches rely on random or predefined communication topologies, without evaluating whether the selected links are actually beneficial for collaboration~\cite{shi2023towards, dai2022dispfl, shi2023improving, kalra2023decentralized, wang2023enhancing}. While \cite{jeong2023personalized} attempted to build a dynamic topology based on client similarity, their method only considers output logit similarity and does not account for the underlying data distribution. This highlights the need for more intelligent and adaptive strategies for link selection that can improve collaboration effectiveness and accelerate convergence in environments with heterogeneous client data.

Furthermore, while differential privacy has been widely explored in centralized FL~\cite{noble2022differentially}, a significant gap remains in addressing both data and model privacy in P2P learning settings~\cite{shi2023towards, wang2022accelerating, dai2022dispfl, jeong2023personalized, shi2023improving}. Ensuring the privacy of client data is essential for maintaining trust and security in decentralized frameworks. Several approaches have been proposed to enhance privacy in peer-to-peer learning. \cite{bellet2018personalized} investigated the trade-off between utility and privacy in peer-to-peer FL. \cite{kalra2023decentralized} introduced proxy models to enable efficient communication without a central server and incorporated differential privacy analysis to strengthen privacy guarantees. However, the added noise from differential privacy can degrade model performance, particularly when client data is non-IID, and merging models from divergent distributions may further harm accuracy. More recently, \cite{bayrooti2023differentially} extended DP-SGD to decentralized learning, aiming to reach consensus on model parameters across clients. Yet, this consensus model may underperform in non-IID scenarios where personalized models are more appropriate. Another direction, explored by \cite{wang2023enhancing}, applied gradient encryption to protect privacy and used succinct proofs to verify gradient correctness. Despite its security benefits, this approach introduced computational and communication overhead~\cite{maheri2025telesparse} and did not address data heterogeneity. These limitations highlight the need for future research to focus on privacy-preserving mechanisms that not only safeguard sensitive information but also account for the heterogeneity and personalization needs inherent in P2P learning.

Several recent works~\cite{mondal2022beas, qin2024blockdfl, shayan2020biscotti, arapakis2023p4l, chen2018machine} have proposed frameworks for P2P learning that assume a stronger threat model, addressing both privacy and malicious security concerns. For instance, Biscotti~\cite{shayan2020biscotti} is a fully decentralized P2P system for privacy-preserving and secure federated learning, leveraging blockchain, differential privacy, and m-Krum~\cite{blanchard2017machine} for secure aggregation. However, this framework is limited to IID data and does not address client heterogeneity in the context of privacy preservation and robustness against poisoning attacks. BlockDFL~\cite{qin2024blockdfl} is another ledger-based system that combines gradient compression and median-based testing with m-Krum to adapt privacy mechanisms for data-heterogeneous clients and improve robustness against data poisoning. Nevertheless, it considers only a label-flipping attack~\cite{tolpegin2020data}, which is regarded as one of the least challenging poisoning attacks~\cite{han2024fedsecurity}. Moreover, their evaluation of malicious security is restricted only to IID clients. BEAS~\cite{mondal2022beas} is another blockchain-based framework that integrates differential privacy, gradient pruning, and Foolsgold~\cite{fung2020limitations} with m-Krum to defend against both privacy and security threats. However, it lacks comparative analysis of its privacy-preserving approach and, like BlockDFL, evaluates performance solely under a label-flipping attack. These limitations reveal a significant gap in DFL frameworks: the lack of solutions that simultaneously support data-heterogeneous clients and operate under a strong threat model encompassing both privacy preservation and protection against data poisoning and model poisoning~attacks.
}

\section{Problem Statement}
\label{sec:problem}

\subsection{System Model}

We consider $M$ clients $\{c_i\}_{i=1}^M$. Each client, $c_i$, holds a local dataset including data points and corresponding labels $\{X_i,Y_i\} \sim D_i$ drawn from its personal distribution $D_i$. Each $c_i$ aims to learn personalized parameters $w_i$ to minimize the expected loss over the client's data distribution: 
\begin{equation}
    \label{eq:loss_minimization}
        F_i(w_i) = \mathbb{E}[ \mathcal{L}_{(x,y) \sim D_i} (w_i ; (x,y)) ].
\end{equation}

To find parameters to minimize the expected loss, each client seeks to determine an optimal parameter set, denoted as $w_i^*$, within the parameter space $w_i \in \mathbb{R}^d$. This optimization aims to minimize the expected loss over the client's own local data distribution $\ell (f_i,y)$, represented in Equation \ref{eq:w_star}:

\begin{equation}
    \label{eq:w_star}
    w_i^* = arg \min_{w_i \in \mathbb{R}^d}  \mathbb{E}_{(x,y)\sim D_i} \ell (f_i (w_i ; x) , y)].
\end{equation}

However, solely using client's local dataset is not sufficient to find generalizable parameters for each client, so local training leads to poor generalization performance~\submit{\cite{li2022learning}}\arxiv{\cite{li2022learning, li2021ditto}}. In order to achieve good generalization, each client is willing to collaborate and share knowledge (\ie their model gradients) with a \emph{subset} of other clients. We note that the goal of the learning is not to reach a ``global consensus'' model, but for each client to obtain personalized models that perform well on their own data distributions.

\subsection{Threat Model}
\label{sec:threat_model}

We consider both privacy and security threats in a P2P setting without a trusted third party or a centralized server.

From a privacy perspective, we assume \textit{semi-honest} clients, as in prior work~\cite{mondal2022beas, shayan2020biscotti}, who may attempt to infer sensitive information from other clients’ private training datasets, but they do not deviate from the training protocol. We assume that clients communicate exclusively over encrypted channels, preventing third-party adversaries from eavesdropping or tampering with the communication. Additionally, clients do not share data directly with others in the network. To mitigate the risk of sensitive information leakage through model updates and to ensure a differential privacy guarantee, each client is assigned a fixed privacy budget in advance.

From a security standpoint, we consider adversaries in a \textit{semi-honest} setting, where clients can deviate from the protocol specification by contributing malicious updates designed to degrade the performance or integrity of the aggregated model. In particular, our threat model assumes the adversary is capable of orchestrating Sybil attacks\arxiv{~\cite{douceur2002sybil}}, thereby controlling multiple peers within the network. However, consistent with prior literature~\cite{shayan2020biscotti, arapakis2023p4l}, we constrain the adversary's stake to no more than $30\%$ of the total participants in the network.
Specifically, our threat model includes prevalent attack scenarios investigated in recent privacy-preserving and peer-to-peer FL frameworks~\cite{mondal2022beas, shayan2020biscotti, arapakis2023p4l, qin2024blockdfl}. These include \lf\arxiv{~\cite{tolpegin2020data}}, \byz\arxiv{~\cite{lin2019free}}, \byzr\arxiv{~\cite{lin2019free}}, and \byzf\arxiv{~\cite{xu2022byzantine}}\submit{~\cite{han2024fedsecurity}}. These attacks collectively represent critical and realistic threats faced by distributed machine learning protocols, particularly in peer-to-peer and privacy-preserving contexts.
In our \lf setting, client data is poisoned by flipping \textit{all labels} as in~\cite{tolpegin2020data}. Byzantine attacks corrupt local model updates by setting weights to zero, random values, or flipped values using $w_g + (w_g - w_l)$, where $w_g$ is the global model, and $w_l$ is the real local model. Additionally, we assume an attacker may have knowledge of the system design, particularly the group formation phase (see Section~\ref{subsection:similar_clients}).

\section{The Design of P4}
\label{sec:design}

\arxiv{In this section, we describe the design of \name in detail. Section~\ref{sec:overview} outlines key challenges and provides an overview of the framework. Section~\ref{subsection:similar_clients} describes how clients form groups based on model weight similarity, while Section~\ref{subsection:private_train} explains the collaborative training algorithm inside each group based on local and proxy models with differential privacy. Next, Section~\ref{subsection:dp} defines the privacy goals and formal guarantees. Finally, Section~\ref{subsection:malicious_security} introduces a combined defense against poisoning attacks based on knowledge distillation, anomaly detection, and \mk.}

\subsection{Overview}
\label{sec:overview}

\name aims to provide a personalized model for each client under DP guarantees and robustness to poisoning attacks, while remaining suitable for deployment on resource-constrained IoT devices. Based on the problem statement in Section~\ref{sec:problem}, we identify four key challenges.

The first challenge is achieving effective client personalization while still allowing clients to benefit from shared knowledge in a system without a central server. This requires a clustering mechanism to group clients with similar data distributions. However, enabling such collaboration introduces the second challenge: maintaining data privacy. Any knowledge exchanged between clients to enhance local model training must not violate differential privacy. Following prior work, this can be achieved by adding Gaussian noise to shared gradients, but this comes at the cost of performance degradation. In a fully decentralized setting, fewer clients can participate in collaborative training compared to centralized federated learning, making the impact of noise more severe, as demonstrated in Section~\ref{sec:privacy_utility}. The third challenge stems from the limited computational and communication resources available to clients. The learning algorithm must converge efficiently within a limited number of iterations, and the knowledge-sharing method should minimize both computation and communication overhead. Finally, malicious clients affected by poisoning attacks can degrade clients' model performance. The system must incorporate an effective defense mechanism that aligns with privacy and personalization requirements without introducing excessive computational~costs.

\begin{figure}[t!]
    \centering
    \includegraphics[width=\linewidth]{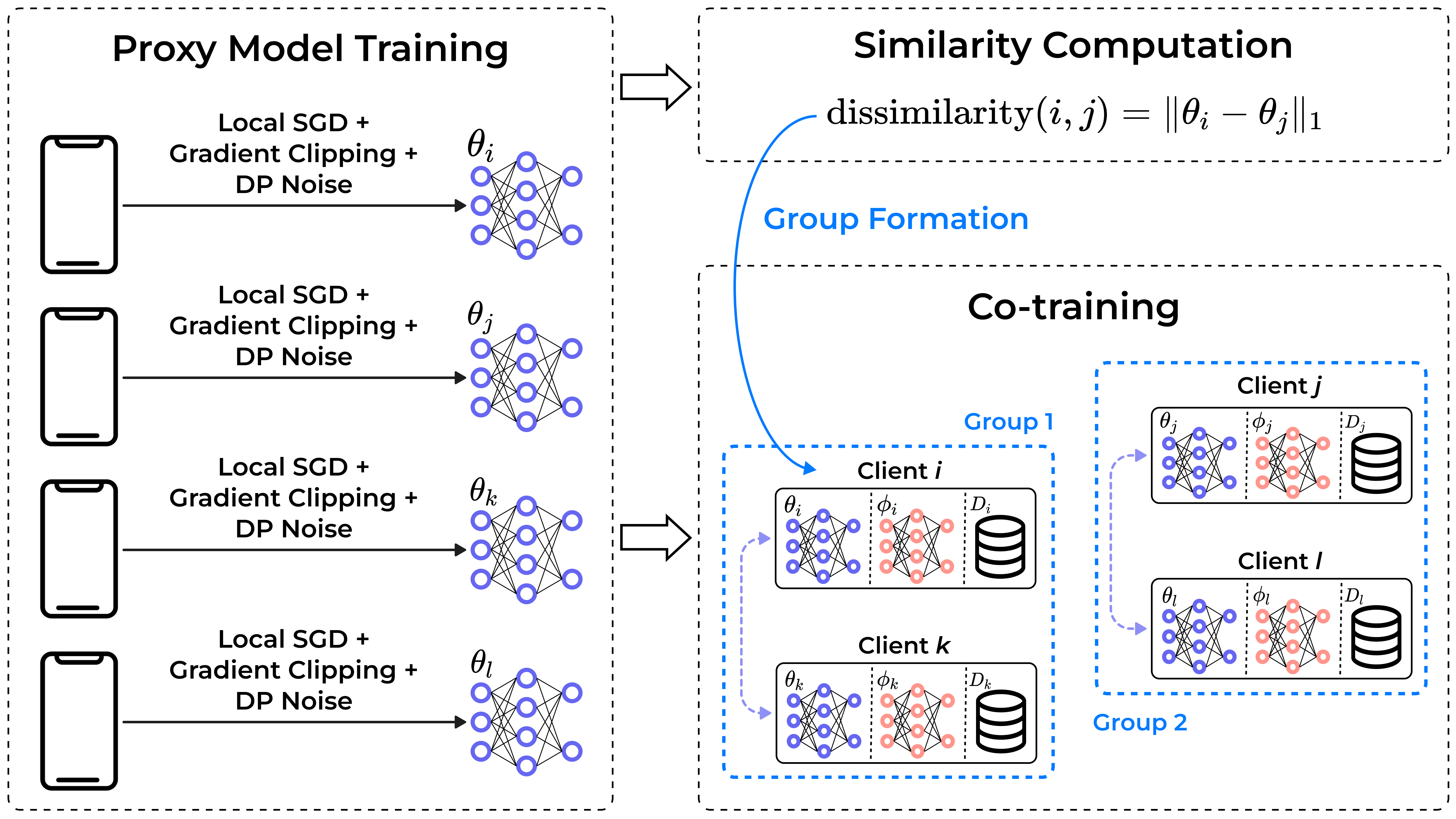}
    \vspace{-0.5cm}
    \caption{The overall design of the \name approach. Clients employ \textit{local} ($\phi$) and \textit{proxy} ($\theta$) models, aggregating only the \textit{proxy} model. Groups are formed based on the $\ell_1$-norm dissimilarity of their model weights. Within each group, updates are exchanged via one client acting as an \textit{aggregator}, which may change during training to balance communication overhead. After receiving the aggregated model, clients perform local training for personalization.}
    \label{fig:overview}
\end{figure}

Figure~\ref{fig:overview} illustrates the overall design of the \name approach, which addresses these challenges. It operates in two phases: \textit{group formation} and \textit{co-training}. During group formation, clients privately identify peers with similar data distributions and form groups accordingly. In the co-training phase, clients train their \textit{local} and \textit{proxy} models and share locally computed data gradients with added noise only within their group. A group aggregator then filters out adversarial model updates, aggregates the remaining gradients, and sends the resulting global model update back to the clients. The following sections provide a more detailed explanation of the system. A table with all the notation is provided in Appendix~\ref{apdx:notations}\submit{ in full paper~\cite{p4_full}}.

\subsection{Phase 1: Group Formation}
\label{subsection:similar_clients}

In federated learning, similar data distributions among clients play an especially important role for model convergence\arxiv{~\cite{dandi2022implicit, karimireddy2020scaffold, huang2024federated}}. Selecting groups of clients for co-training without considering the underlying data heterogeneity---such as through random clustering---can lead to performance worse than standalone local training~\submit{\cite{li2022towards}}\arxiv{\cite{zec2023private, li2022towards}}. Therefore, identifying clients with similar data distributions is critical for effective collaborative training.

A direct measure of data distribution similarity across decentralized clients can be challenging due to privacy constraints. Inspired by prior works~\submit{\cite{ghosh2020efficient}}\arxiv{\cite{ghosh2020efficient, sattler2020clustered}}, we use the similarity of client model parameters as an effective proxy. Specifically, we measure dissimilarity using the $\ell_1$-norm between model~weights:

\begin{equation}
\label{eq:l1_similarity}
\text{dissimilarity}(i,j) = \lVert \mathbf{w}_i - \mathbf{w}_j \rVert_1.
\end{equation}

The intuition behind using model weight similarity as a proxy stems from the observation that similar model parameters imply similar gradient behaviors. Specifically, consider two clients, $c_i$ and $c_j$, each with personalized model parameters $w_i$ and $w_j$, respectively. The gradient of the loss function for client $c_j$, evaluated at client $c_i$'s parameters, can be approximated using a first-order Taylor expansion around $w_j$:

\vspace{-0.1cm}

\begin{align}
\nabla_{w_i} F_j(w_i) = \nabla_{w_j} F_j(w_j)  &+ \nabla_{w_j}^2 F_j(w_j)(w_i - w_j) \nonumber \\ \
& + O(|w_i - w_j|^2).
\end{align}

The Hessian matrix $\nabla_{w_j}^2 F_j(w_j)$ characterizes the curvature of the loss landscape around $w_j$, assigning importance to each parameter dimension. Thus, the approximation error incurred by client $c_i$ when leveraging gradients from client $c_j$ is closely related to the parameter distance:

\vspace{-0.1cm}

\begin{equation}
\text{Error} \approx \frac{1}{2}(w_i - w_j)^T \nabla_{w_j}^2 F_j(w_j) (w_i - w_j).
\end{equation}

Minimizing the distance between model parameters directly reduces the approximation error, thereby improving the accuracy and effectiveness of collaborative training. To ensure computational efficiency---crucial for resource-constrained decentralized IoT environments---our method avoids explicit Hessian computation and instead focuses on minimizing model parameter differences. This theoretical foundation motivates our client similarity metric and underscores the importance of parameter proximity for effective collaborative training and personalization in decentralized settings.

To effectively capture meaningful model differences reflecting local data distributions, each client initially performs minimal local training---specifically, one epoch over its local dataset---to ensure parameter vectors reflect local data characteristics. This local training phase incorporates gradient clipping and differential privacy noise\arxiv{, as detailed in Section~\ref{subsection:dp}}. Moreover, since models are only partially trained and have not converged, they primarily encode coarse-grained information about the general direction of the data rather than specific details, which significantly reduces vulnerability to attacks such as gradient leakage or data reconstruction. Empirical results (Section~\ref{sec:privacy_utility}) and theoretical analysis (Appendix~\ref{sec:one_epoch_distinguish}\submit{ in the full paper~\cite{p4_full}}) further confirm that even this minimal training is sufficient to distinguish between substantially different data distributions without incurring major privacy risks.

Clients are grouped based on a dissimilarity metric. Let $M$ denote the total number of clients in the network. The grouping structure is represented by a binary matrix $G \in {0,1}^{M \times M}$, where $G_{ij} = 1$ indicates that clients $i$ and $j$ belong to the same group, and $G_{ij} = 0$ otherwise. The grouping objective is then defined as follows:

\begin{equation}
    \underset{G}{min} \sum_{i=1}^M \sum_{j=1}^M  \mathbb{I}[G(i,j)=1] \lVert \mathbf{w}_i - \mathbf{w}_j \rVert_1,
\end{equation}

such that:
\begin{equation}
    \sum_{j=1}^{M} G_{i,j} = V \quad \text{for } i = 1,2,\ldots,M
\end{equation}

where $G$ is the collaboration graph (symmetric matrix) and $V$ defines a group size (due to limited computation and communication of each client). We note that \name enforces mutual collaborative training, implying that communication links between clients are bidirectional, and the matrix $G$ should be symmetric.

To construct the collaboration graph, $G$, in a decentralized manner, we employ a greedy approach to identify the $V$ most similar models for each client (see \arxiv{Algorithm~\ref{alg:group_formation}}\submit{a pseudocode in the full paper~\cite{p4_full}}). First, each client computes its similarity with $H$ randomly selected clients ($H \ll V$) using the model similarity metric in Equation \ref{eq:l1_similarity}\arxiv{ (lines 1-7)}. Next, if two clients mutually select each other as their most similar among the $H$ samples, they form a two-member group\arxiv{ (lines 8-16)}. Unassigned clients, such as client $t$, join their most similar ungrouped client $k$, regardless of whether $t$ is also $k$'s most similar sample\arxiv{ (line 17)}. Any remaining ungrouped clients are paired randomly, ensuring all clients are part of exclusive two-member groups\arxiv{ (line 18)}. Within each group, clients share their computed similarities. The group then estimates its similarity with other groups as the maximum similarity between any of its members and those in the target group\arxiv{ (line 19)}. The process iterates from the second step until the desired group size is reached. Privacy guarantees of the group formation phase are explained in \submit{the full paper~\cite{p4_full}}\arxiv{Section~\ref{subsection:dp}}.

\arxiv{\SetAlFnt{\small} 
\SetAlCapFnt{\small}
\SetAlCapNameFnt{\small}

\begin{algorithm}[t!]
\caption{Client Grouping Based on L1-Norm Similarity}
\label{alg:group_formation}

\DontPrintSemicolon
\SetAlgoNlRelativeSize{-2}
\SetAlCapNameFnt{\small}
\SetNlSty{textbf}{}{}  

\KwIn{Set of clients $M$, group size $V$, random subset size $H$.}

\KwOut{A binary matrix $G$ with client grouping based on $\ell_1$-norm.}

\tcp{Step 1: Compute l1-norm similarity among clients}

$\text{\textit{mcs}} \gets \emptyset$   \texttt{//} Dictionary: client → list of similarity tuples\;
\ForEach{client $i \in M$}{
    \ForEach{client $j \in $ random subset $S \subset M \setminus \{i\}, |S| = H$}{
        \If{$(i, j) \notin \text{\textit{mcs}}$}{
            $\text{\textit{mcs}}[i] \leftarrow \text{\textit{mcs}}[i] \cup \{(j, \text{dissimilarity}(\mathbf{x}_i, \mathbf{x}_j))\}$\;
            
            $\text{\textit{mcs}}[j] \leftarrow \text{\textit{mcs}}[j] \cup \{(i, \text{dissimilarity}(\mathbf{x}_j, \mathbf{x}_i))\}$\;
        }
    }
}
Sort each $\text{\textit{mcs}}[i]$ in descending order of similarity\;

\tcp{Step 2: Create pairs of mutual similar clients}

$\text{\textit{similar\_clients}}[i] \leftarrow \emptyset$ for all $i \in M$\;
\ForEach{client $i \in M$}{
    \lIf{$\text{\textit{similar\_clients}}[i] \neq \emptyset$}{Continue}
    
    $k \gets mcs[i][0][0]$ \tcp*{Most similar client}
    \If{$i = mcs[k][0][0]$ and $\text{\textit{similar\_clients}}[k] = \emptyset$}{
        $\text{\textit{similar\_clients}}[i] \leftarrow \text{\textit{similar\_clients}}[i] \cup \{k\}$\;
        
        $\text{\textit{similar\_clients}}[k] \leftarrow \text{\textit{similar\_clients}}[k] \cup \{i\}$\;
        
        Remove $(k, \cdot)$ from $\text{\textit{mcs}}[i]$\;
        
        Remove $(i, \cdot)$ from $\text{\textit{mcs}}[k]$\;
    }
}

Unassigned clients join their most similar unassigned client\;
Remaining clients create pairs randomly\;

\tcp{Step 3: Combine groups}

Merge groups based on similarity until reaching $V$ group size\;
$G \gets$ (final merged groups)\;

\Return $G$\;

\end{algorithm}
}

\subsection{Phase 2: Co-training}
\label{subsection:private_train}

After group formation, clients within a group engage in collaborative training. To mitigate data privacy leakage, differential privacy noise must be applied. However, in a fully decentralized system, DP noise can more severely impact accuracy compared to centralized training due to the limited number of clients sharing knowledge~\submit{\cite{lalitha2019peer}}\arxiv{\cite{10.5555/3692070.3692117,ijcai2024p635}}, constrained by their computational and communication power. Thus, an efficient algorithm to aggregate knowledge between clients is needed to ensure fast convergence and personalization for each client.

In \name, we use transformed features from a handcrafted layer~\cite{tramer2020differentially} as input to the client’s neural network. These features enhance convergence even in the presence of DP noise and enable strong performance with a shallow neural network (see Section~\ref{sec:privacy_utility}). The transformed features feed into two models: a \textit{local} model, $f_{\phi_i}$, and a \textit{proxy} model, $f_{\theta_i}$ (see \submit{a pseudocode in the full paper~\cite{p4_full}}\arxiv{Algorithm~\ref{alg:private_training}}). Clients share only the proxy model within their group. During each training round, both models train on local data $D_i$, but the proxy model’s gradients are clipped and DP noise is added. Since the proxy model undergoes DP training, sharing its gradient updates $\Delta \theta_i$ does not violate DP guarantees as explained in \submit{the full paper~\cite{p4_full}}\arxiv{Section \ref{subsection:dp}}. Clients exchange these proxy model gradients with their group members, incorporating external knowledge into $f_{\theta_i}$ while preserving local data privacy.

To train the local model and the proxy model together, inspired by \cite{kalra2023decentralized}, we apply knowledge distillation between these two models\arxiv{~\cite{zhang2018deep}}. More specifically, training of the local model will be done by classification loss (Equation~\ref{eq:proxy_classification}) and KL divergence loss (Equation~\ref{eq:proxy_kl}):

\begin{equation}
    \label{eq:proxy_classification}
    \mathbf{L}_{CE} (f_{\theta_i}) = \mathbb{E}_{(x,y) \sim D_i} CE[f_{\theta_i}(x) \parallel y],
\end{equation}

where $CE$ denotes the cross-entropy loss, computed between the proxy model’s output and the ground-truth label $y$ corresponding to input $x$.

In addition, to distill knowledge from the local model to the proxy model, we use the Kullback–Leibler divergence described in Equation~\ref{eq:proxy_kl} to make the proxy model output closer to the local model output:

\begin{equation}
    \label{eq:proxy_kl}
    \mathbf{L}_{KL} (f_{\theta_i} ; f_{\phi_i}) = \mathbb{E}_{(x,y) \sim D_i} KL[f_{\theta_i}(x) \parallel f_{\phi_i}(x)].
\end{equation}

Therefore, the proxy model learns from both the local model and the local data by combining the classification loss and the KL divergence loss shown in Equation~\ref{eq:proxy_loss}. This way, the proxy model learns to correctly predict the true label of training instances as well as to match the probability estimate of its local model:

\begin{equation}
    \label{eq:proxy_loss}
    \mathbf{L}_{\theta_i} = (1-\alpha) \cdot \mathbf{L}_{CE} (f_{\theta_i}) + \alpha \cdot \mathbf{L}_{KL} (f_{\theta_i} ; f_{\phi_i}).
\end{equation}

In the combination of the losses, $\alpha \in [0,1]$ balances between losses. As it increases, the proxy model will use more information from the local model.

The objective of the local model is shown in Equation~\ref{eq:private_loss}:

\begin{equation}
    \label{eq:private_loss}
    \mathbf{L}_{\phi_i} = (1-\beta) \cdot \mathbf{L}_{CE} (f_{\phi_i}) + \beta \cdot \mathbf{L}_{KL}(f_{\phi_i} ; f_{\theta_i}).
\end{equation}

As $\beta \in [0,1]$ increases, the local model will rely less on its local data and more on the data from other groups, which is in the proxy model. If $\beta$ is zero, it results in a totally personalized model that is only trained on the local training data.

With this architecture, the local model $f_{\phi_i}$ can extract personalized model weights for the client without dealing with DP noise, leading to improved performance. In other words, by decoupling \textit{local} models from the differential privacy noise, we aim to enhance model performance, resulting in models that better align with the specific data characteristics of individual clients. Besides that, the parameters of the proxy model $f_{\theta_i}$ could be aggregated with neighbor clients to benefit from their knowledge under DP guarantees.

We note that it is not necessary for each member of a group to receive model updates $\Delta \theta_j$ from all other members and update its proxy model $\theta_i$. Instead, every few rounds, an aggregator can be selected periodically from within the group. Therefore, other clients within the group only send their model updates $\Delta \theta_j$ to the aggregator client and subsequently receive the aggregated model $\theta$ from that client, which is used for the next round of co-training.

\arxiv{\begin{algorithm}[t!]
\caption{Co-training Inside One Group}
\label{alg:private_training}

\DontPrintSemicolon
\SetAlgoNlRelativeSize{-2}  
\SetNlSty{textbf}{}{}

\KwIn{Proxy $\theta_i^{(0)}$ and local $\phi_i^{(0)}$ parameters after phase 1, distillation weights $\alpha, \beta \in (0,1)$, learning rates $\eta_l, \eta_g > 0$, client subsampling $l$, data subsampling $s$, local dataset size $R$.}

\KwOut{Proxy $\theta_i^{(T)}$ and local $\phi_i^{(T)}$ models for each client co-trained inside a group under DP guarantees.}

$\hat{M} \gets$ (one group clients from $G$)\;

\tcp{In parallel}
\For{round $t = 0, \dots, T-1$ at client $i \in C^t \subset [\hat{M}]$ of size $\lfloor l\hat{M} \rfloor$}{
    \For{local optimization step $k = 0, \dots, K-1$}{
        Sample $S_i^{k} \subset D_i$ of size $\lfloor sR \rfloor$\;
        
        Update local and proxy models:\;
        \quad $\theta_i^{(t)} \gets \theta_i^{(t)} - \eta_l \nabla \tilde{\mathcal{L}}_{\theta_i}(S_i^{k})$ \tcp*{DP update}
        \quad $\phi_i^{(t)} \gets \phi_i^{(t)} - \eta_l \nabla \hat{\mathcal{L}}_{\phi_i}(S_i^{k})$ \tcp*{Non-DP update}
    }
    
    $\phi_i^{(t+1)} \gets \phi_i^{(t)}$\;

    $\Delta \theta_i^{(t)} \gets \theta_i^{(t)} - \theta^{(t - 1)}$\;

    \If{client i is a group aggregator}{
        $\Delta \theta^{(t)} \gets \frac{1}{l\hat{M}} \sum_{j \in C^t} \Delta \theta_j^{(t)}$\;

        $\theta^{(t)} \gets \theta^{(t - 1)} + \eta_g \Delta \theta^{(t)}$\;

        Send $\theta^{(t)}$ to clients $j \in C^t$\;
    }
    \Else{
        Send $\Delta \theta_i^{(t)}$ to a group aggregator\;
        
        Receive $\theta^{(t)}$ from a group aggregator\;
    }
    
    Update proxy model: \quad $\theta_i^{(t+1)} \gets \theta^{(t)}$\;
}

\Return $\theta_i^{(T)}$, $\phi_i^{(T)}$\;

\end{algorithm}
}

\arxiv{\subsection{Differential Privacy}
\label{subsection:dp}

\paragraph{Privacy goal} We aim to control the information leakage from individual data of each client $D_i$ in the shared model updates $\Delta \theta_i$. To achieve this, we adapt a differential privacy algorithm for heterogeneous data proposed by~\cite{noble2022differentially} to the peer-to-peer learning setting, and ensure the privacy guarantees on top of their theoretical proof. We focus on \textit{record-level} DP with respect to the joint dataset $D$, where $D$ and $D^{\prime}$ are neighboring datasets if they differ by at most one record (denoted by $\|D - D^{\prime}\|\le1$). We want to ensure privacy (i) towards a third-party observing the final model and (ii) an honest-but-curious aggregator in each group (similar to an honest-but-curious server in~\cite{noble2022differentially}). We set the DP budget for the whole training, denoted by $(\epsilon,\delta)$, in advance before group formation (phase 1) and use the same clipping and Gaussian noise as in the original work to achieve this budget.

\paragraph{Privacy guarantees} To ensure privacy during co-training (see Algorithm~\ref{alg:private_training}), each gradient of a proxy model of $i$-th client is divided by its norm as described in Equation~\ref{eq:clip} and then the noise is added to the gradient as shown in Equation~\ref{eq:noise} (line 6 in Algorithm~\ref{alg:private_training}):

\begin{equation}
    \label{eq:clip}
    \Tilde{g}_{ij} = \frac{g_{ij}}{\max(1,\| g_{ij} \|_2 / \mathcal{C})},
\end{equation}

\begin{equation}
    \label{eq:noise}
    \Tilde{H_i} = \frac{1}{sR} \Sigma_{j\in S_i} \Tilde{g}_{ij} + \frac{2\mathcal{C}}{sR} \mathcal{N}(0,\sigma_g^2).
\end{equation}

The noise standard deviation $\sigma_g$ in Equation~\ref{eq:noise}, ensuring the specified privacy budget ($\epsilon$) given the clipping norm defined in Equation~\ref{eq:clip}, is computed based on the theoretical upper bound adapted from Theorem 4.1 of~\cite{noble2022differentially}, as follows:

\begin{equation}
    \label{eq:dp_noise}
    \sigma_g = \Omega( \frac{s \sqrt{lTK \log(\frac{2Tl}{\delta}) \log(\frac{2}{\delta})}} {\epsilon \sqrt{M^\prime}}).
\end{equation}

It guarantees that under the subsampled Gaussian mechanism~\cite{wang2019subsampled} and composition accounting via Rényi DP~\cite{mironov2017renyi}, one can ensure (i) $(\mathcal{O}(\epsilon), \delta)$-differential privacy towards a third party observing the global model (\eg other clients that receive the global model after aggregation), and (ii) $(\mathcal{O}(\epsilon_s), \delta_s)$-DP towards an honest-but-curious aggregator receiving per-round gradients, where $\epsilon_s = \epsilon \sqrt{M^{'}/l}$ and $\delta_s = \frac{\delta}{2} (\frac{1}{l} + 1)$. We follow their privacy accounting strategy matched to the peer-to-peer setting. Since in P2P learning individual gradients are directly shared among clients rather than being aggregated, we set $M^\prime$ to 1. To amplify privacy according to~\submit{\cite{noble2022differentially}}\arxiv{\cite{kasiviswanathan2011can, noble2022differentially}}, client subsampling $l$ and data subsampling $s$ are used (lines 2 and 4 in Algorithm~\ref{alg:private_training}). Based on the Equation \ref{eq:noise}, once the target privacy budget $\epsilon$ and $\delta$ are fixed, one can compute a valid set of values for $K$, $l$, and $s$ to satisfy the guarantee. We specify the chosen values for these parameters in Section~\ref{sec:privacy_utility} and Appendix~\ref{apdx:hyperparams}.
}

\submit{\subsection{Defense against Poisoning Attacks}
\label{subsection:malicious_security}

In \name, we account for a wide range of poisoning attacks, defined in Section~\ref{sec:threat_model}. To mitigate these threats while maintaining privacy and utility amplification, \name employs a combination of \mk~\cite{blanchard2017machine}, \anom~\cite{han2023kick}, and knowledge distillation, which utilizes client similarity within each group to filter out malicious updates. To the best of our knowledge, \name incorporates a unique set of defense strategies not used in prior P2P work, simultaneously enhancing the privacy-utility trade-off and improving robustness against poisoning attacks under heterogeneous client data distributions. Note that \mk and \anom are existing defenses used during gradient aggregation; for simplicity, we refer to their combination as ``secure aggregation.'' The following subsections detail the rationale for each defense strategy, while Section~\ref{sec:tolerating_attacks} provides an empirical evaluation of the proposed defense mechanism.\submit{ A more detailed description of the proposed defense can be found in the full paper~\cite{p4_full}.}

\subsubsection{Data Poisoning Tolerance}

\name demonstrates inherent resilience against data poisoning attacks---particularly \lf~\cite{han2024fedsecurity}---through its combination of proxy-based knowledge distillation and client clustering, as introduced in Sections~\ref{subsection:similar_clients} and \ref{subsection:private_train}. Central to this defense is the use of knowledge distillation, which enhances the privacy-utility trade-off by aligning the local model $f_{\phi_i}$ with the proxy model $f_{\theta_i}$ via a KL-divergence loss term. When the KL term is incorporated into the loss functions $\mathbf{L}_{\theta_i}$ and $\mathbf{L}_{\phi_i}$ (Equations~\ref{eq:proxy_loss} and \ref{eq:private_loss}), it enforces alignment between the softmax outputs of the \textit{clean} local model $f_{\phi_i}$ and the \textit{poisoned} proxy model $f_{\theta_i}$. Similar to \texttt{LeadFL}~\cite{zhu2023leadfl}, which introduces an additional regularization term in the loss function to mitigate the effects of poisoned gradients, the KL-divergence term in \name's loss function also serves as a regularization mechanism.

\subsubsection{Model Poisoning Tolerance}

While an unmodified \name effectively mitigates data poisoning attacks, it is less robust against model poisoning attacks (see Section~\ref{sec:tolerating_attacks}). The primary challenge lies in the significant changes in client gradients (\eg setting them to zero or random values), which can disrupt simple gradient averaging on the aggregator. To mitigate Byzantine attacks, \name integrates a robust aggregation method called \mk~\cite{blanchard2017machine}. The integration of \mk with \name’s client clustering mechanism substantially enhances its ability to detect and filter out malicious updates (see Section~\ref{sec:tolerating_attacks}).

\submit{
\begin{figure*}[t!]
    \centering
    \subfigure[]{\includegraphics[width=0.26\textwidth]{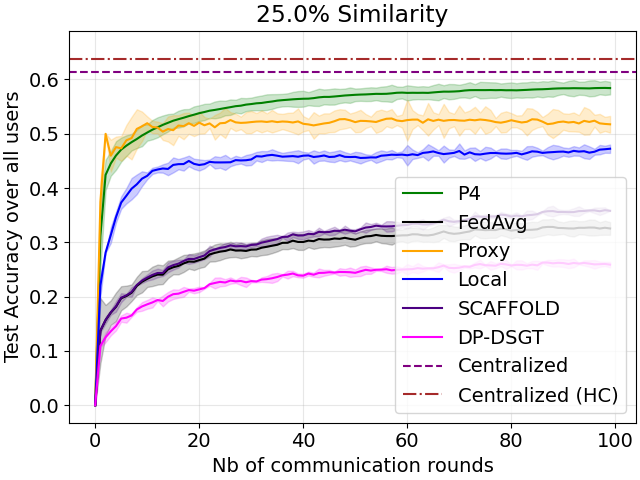}}
    \hspace{0.75cm}
    \subfigure[]{\includegraphics[width=0.26\textwidth]{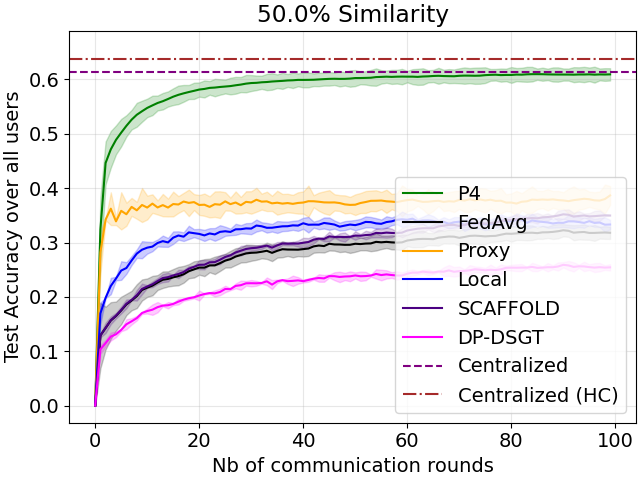}} 
    \hspace{0.75cm}
    \subfigure[]{\includegraphics[width=0.26\textwidth]{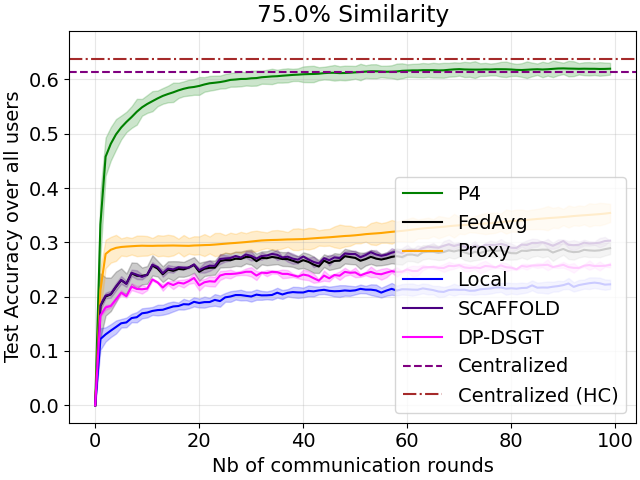}}
    \vspace{-0.2cm}
    \caption{Test accuracy of a linear model on \cifar with $\epsilon=15$ and alpha-based setting: (a) $\gamma=25\%$ (b) $\gamma=50\%$ (c) $\gamma=75\%$. For each client, $\gamma\%$ of the data is sampled IID from all classes, while the remaining $1-\gamma\%$ comes from a single dominant class.}
    \vspace{-0.25cm}
    \label{fig:diff_sim}
\end{figure*}

\begin{figure*}[t!]
    \centering
    \subfigure[]{\includegraphics[width=0.26\textwidth]{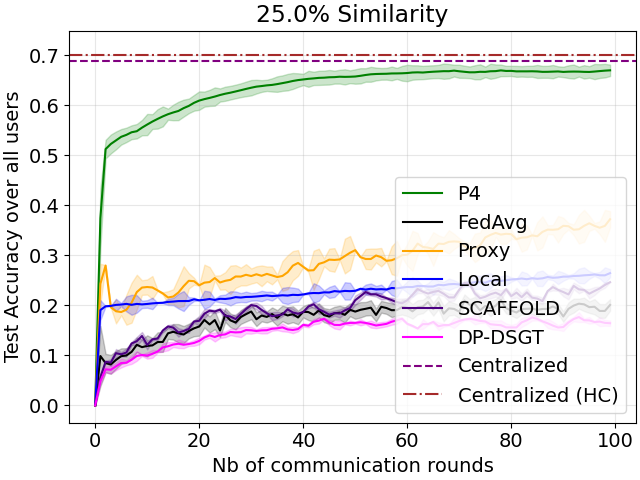}}
    \hspace{0.75cm}
    \subfigure[]{\includegraphics[width=0.26\textwidth]{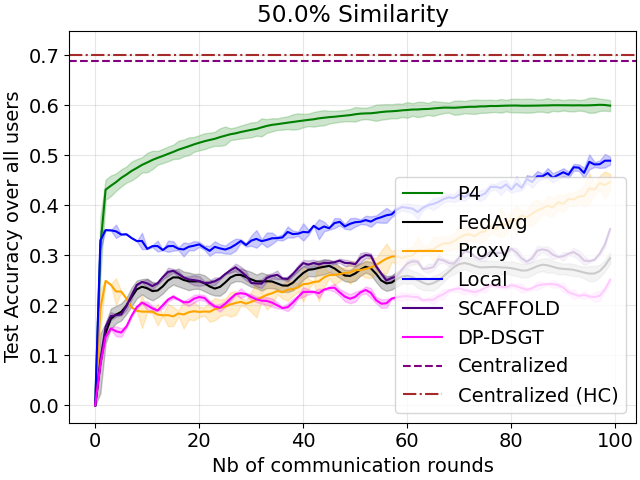}}
    \hspace{0.75cm}
    \subfigure[]{\includegraphics[width=0.26\textwidth]{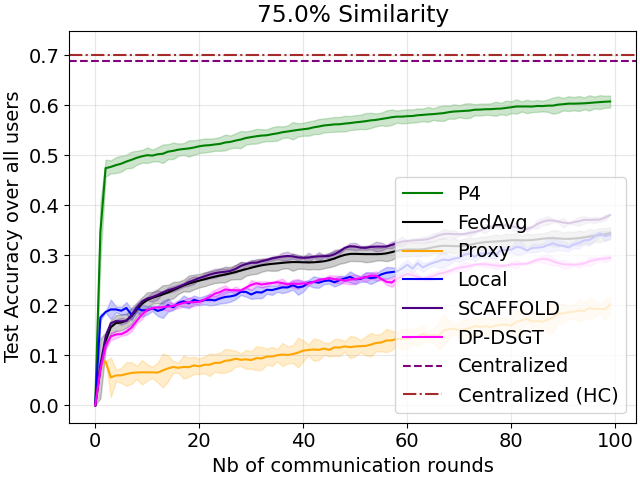}}
    \vspace{-0.2cm}
    \caption{Test accuracy of CNN on \cifar with $\epsilon=15$ and alpha-based non-IID setting: (a) $\gamma=25\%$ (b) $\gamma=50\%$ (c) $\gamma=75\%$. For each client, $\gamma\%$ of the data is sampled IID from all classes, while the remaining $1-\gamma\%$ comes from a single dominant class.}
    \vspace{-0.25cm}
    \label{fig:diff_sim_cnn}
\end{figure*}
}

\subsubsection{Anomaly Detection}  

To strengthen \name's resilience against anomalous model updates, particularly in the \byzr attack, we incorporate \texttt{anomaly-detection} \cite{han2023kick}. Our empirical evaluation in the full version of the paper~\cite{p4_full} shows that while \mk struggles against \byzr, such a simple and comparably fast defense as \texttt{anomaly-detection}~\cite{han2023kick} is highly effective. It leverages the $3\sigma$ rule applied to an $\ell_2$ score during cross-client checks, making it particularly suitable as an addition to \mk and \name's clustering approach. Since clients within each group are expected to have high model weight similarity after clustering, anomalous updates can be easily identified and filtered during training. Moreover, in cases where \texttt{anomaly-detection} is less effective, such as in \lf, it does not degrade performance\arxiv{(see Figure~\ref{fig:all_attacks}(a))}. Thus, we integrate \texttt{anomaly-detection} as the first layer of defense in \name, complementing \mk and client self-defense to enhance overall robustness.
}\arxiv{\subsection{Defense against Poisoning Attacks}
\label{subsection:malicious_security}

In \name, we account for a wide range of poisoning attacks, defined in Section~\ref{sec:threat_model}. To mitigate these threats while maintaining privacy and utility amplification, \name employs a combination of \mk~\cite{blanchard2017machine}, \anom~\cite{han2023kick}, and knowledge distillation, which utilizes client similarity within each group to filter out malicious updates. To the best of our knowledge, \name incorporates a unique set of defense strategies not used in prior P2P work, simultaneously enhancing the privacy-utility trade-off and improving robustness against poisoning attacks under heterogeneous client data distributions. Note that \mk and \anom are existing defenses used during gradient aggregation; for simplicity, we refer to their combination as ''secure aggregation.'' The rationale behind each defense strategy is discussed in the following subsections.

\subsubsection{Data Poisoning Tolerance}

We use a \lf attack to implement targeted data poisoning, which is particularly prevalent in the DFL security literature~\submit{\cite{fang2024byzantine, heydaribeni2023surefed}}\arxiv{\cite{fang2024byzantine, heydaribeni2023surefed, sun2024byzantine, he2022byzantine}}. Given a source class $c_{src}$ and a target class $c_{target}$ from $\mathbf{C}$, each malicious client $i \in M$ modifies their dataset $D_i$ as follows: For all instances in $D_i$ whose class is $c_{src}$, change their class to $c_{target}$. In our setting, client data is poisoned by flipping \textit{all labels} (\eg for \cifar from $c_{src}$ to $9-c_{src}$ as in~\cite{tolpegin2020data}). The goal of the attack is to make the proxy model of each client $f_{\theta_i}$ more likely to misclassify $c_{src}$ images as $c_{target}$ images at test time.

Interestingly, a proxy-based knowledge distillation combined with \name's client clustering presented in Sections~\ref{subsection:similar_clients} and \ref{subsection:private_train} shows inherent resilience against data poisoning attacks, particularly \lf. While Section~\ref{sec:tolerating_attacks} provides an empirical evaluation (see Figures~\ref{fig:all_attacks}(a) and \ref{fig:ideal_vs_p4_defense}), we present an intuition for this tolerance here.

\name enhances the privacy-utility trade-off by employing knowledge distillation from a local model to a proxy model (Eq.~\ref{eq:proxy_kl}) that can be formulated as the KL-divergence between the softmax outputs:

\vspace{-0.3cm}

$$\mathbf{L}_{KL} (f_{\theta_i} ; f_{\phi_i}) = \mathbb{E}_{(x,y) \sim D_i} KL[\text{softmax}(\frac{z_{\theta_i}}{T}) \parallel \text{softmax}(\frac{z_{\phi_i}}{T})]$$

\noindent where $z_{\theta_i}$ and $z_{\phi_i}$ are the logits of proxy and local models, respectively, and $T$ is a temperature that is normally set to~1. From a malicious security perspective, this knowledge distillation mechanism acts as a form of client \textit{self-defense} against poisoning attacks. Similar to \texttt{LeadFL}~\cite{zhu2023leadfl}, which introduces an additional regularization term in the loss function to mitigate the effects of poisoned gradients, the KL-divergence term in \name's loss function also serves as a regularization mechanism. This term helps control the influence of poisoned gradients on the final proxy model performance.

When the KL term is incorporated into loss functions $\mathbf{L}_{\theta_i}$ and $\mathbf{L}_{\phi_i}$ (Equations~\ref{eq:proxy_loss} and \ref{eq:private_loss}), it enforces alignment between the softmax outputs of the proxy model $f_{\theta_i}$ and the local model $f_{\phi_i}$. Even if the gradients of the proxy model $f_{\theta_i}$ are poisoned, \textit{the local model $f_{\phi_i}$ remains clean}. Consequently, the KL alignment between these two models serves as a corrective mechanism, cleansing the proxy model $f_{\theta_i}$ from the effects of poisoning. This client self-defense strategy is particularly effective against \lf, as this attack aims to alter the softmax output of the proxy model $f_{\theta_i}$ by modifying labels. However, the KL alignment with the softmax output of the local model $f_{\phi_i}$ counteracts these changes from the client side, mitigating the impact of the attack.

Nevertheless, due to high data heterogeneity across clients, using a proxy mechanism alone does not ensure robustness under arbitrary clustering settings. This is because when malicious updates poison the proxy models, benign clients attempt to align their poisoned proxy models’ softmax outputs with their \textit{local} models, guided by their local data distribution. However, when these partially cleansed proxy models are averaged under random clustering and client heterogeneity, the differences in benign clients’ data distributions diminish the overall effectiveness of the cleansing process. This requires a clustering algorithm that groups clients with similar data distributions, such as our group formation algorithm (see Appendix~\ref{apdx:exp2:data_poisoning} for supporting details).

\subsubsection{Model Poisoning Tolerance}

While an unmodified \name effectively mitigates data poisoning attacks, it is less robust against model poisoning attacks (see Section~\ref{sec:tolerating_attacks}). The primary challenge lies in the significant changes in client gradients (\eg setting them to zero or random values), which can disrupt simple gradient averaging on the aggregator. To mitigate Byzantine attacks, \name integrates a robust aggregation method called \mk~\cite{blanchard2017machine}. 

\mk defines a \textit{Krum} aggregation rule $\text{Kr}(w_i, \ldots, w_n)$ as follows. For any $i\neq j$, they denote by $i \rightarrow j$ the fact that $w_j$ belongs to $n - f - 2$ closest vectors to $w_i$, where $n$ is the total number of clients and $f$ is the number of malicious clients. Then, they define for each client $i$, the score $s(i) = \sum_{i \rightarrow j} \| w_i - w_j \|^2$, where the sum runs over $n - f - 2$ closest vectors to $w_i$ based on an Euclidean distance. Finally, $\text{Kr}(w_i, \ldots, w_n) = w_{i_*}$, where $i_*$ refers to the client minimizing the score, $s(i_*) \le s(i)$ for all $i$. An $m$ parameter in \mk defines the number of clients to choose that have minimum~scores.

In this setting, client clustering based on model weight similarity before co-training enhances the ability to detect malicious updates. This is because \name forms groups using the $\ell_1$-norm, which was chosen for its effectiveness in client clustering for privacy-utility amplification. Although the $\ell_1$-norm differs from the $\ell_2$-norm, clients that are close under the $\ell_1$-norm are also close under the $\ell_2$-norm. As a result, benign gradients within a group remain close to one another during training, thereby improving \mk’s ability to identify and filter out outliers introduced by malicious clients. Moreover, the use of knowledge distillation between the local and proxy models helps mitigate cases where \mk fails to filter out malicious updates and a proxy model becomes poisoned. Since the local model remains clean, knowledge distillation helps cleanse the poisoned proxy model. We empirical evaluate an effectiveness of \mk with \name against other defenses in Section~\ref{sec:tolerating_attacks}.

\subsubsection{Anomaly Detection}  

To strengthen \name's resilience against anomalous model updates, particularly in the \byzr attack mentioned in Section~\ref{sec:threat_model}, we incorporate \texttt{anomaly-detection} \cite{han2023kick}. Our empirical evaluation in Section~\ref{sec:tolerating_attacks} shows that while \mk struggles against \byzr, such a simple and comparably fast defense as \texttt{anomaly-detection}~\cite{han2023kick} is highly effective. It leverages the $3\sigma$ rule applied to an $\ell_2$ score during cross-client checks, making it particularly suitable as an addition to \mk and \name's clustering approach. Since clients within each group are expected to have high model weight similarity after clustering, anomalous updates can be easily identified and filtered during training. Moreover, in cases where \texttt{anomaly-detection} is less effective, such as in \lf, it does not degrade performance \arxiv{(see Figure~\ref{fig:all_attacks}(a))}. Thus, we integrate \texttt{anomaly-detection} as the first layer of defense in \name, complementing \mk and client self-defense to enhance overall robustness.
}

\section{Experiments}
\label{sec:experiments}

\subsection{Experimental Setup}
\label{sec:exp_setup}

\subsubsection{Datasets}  
We conduct experiments on three benchmark datasets:  \cifar\arxiv{~\cite{krizhevsky2009learning}}, \cifarext\arxiv{~\cite{krizhevsky2009learning}}, and \femnist\arxiv{~\cite{cohen2017emnist}}.\arxiv{ Following the non-IID setting from FedAvg\arxiv{~\cite{mcmahan2017communication}}, we randomly assign classification tasks and training data to each client. This process generates $M$ clients, each with $R$ samples, split into 80\% training and 20\% testing. Additionally, 20\% of clients are reserved for evaluation and hyperparameter tuning.} Dataset statistics are provided in \arxiv{Table~\ref{tab:datasets}}\submit{Table 1 in the full paper~\cite{p4_full}}.

\arxiv{\begin{table}[t!]
    \caption{Dataset statistics.}
    \vspace{-0.25cm}
    \label{tab:datasets}
    \begin{tabular}{ccccc}
        \toprule
        \textbf{Dataset}   & \textbf{\# classes}   & \textbf{\# clients}   & \makecell[tc]{\textbf{\# samples} \\ \textbf{(per client)}} & \textbf{\# features}   \\ 
        \midrule
        FEMNIST   & 47  & 200 & 300 & $28\times28\times1$ \\ 
        CIFAR-10  & 10  & 260 & 200 & $32\times32\times3$ \\ 
        CIFAR-100 & 100 & 60  & 250 & $32\times32\times3$ \\
        \bottomrule
    \end{tabular}
    \vspace{-0.5cm}
\end{table}
}

\subsubsection{Non-IID Setting}  
\submit{To generate non-IID tasks and analyze the impact of data heterogeneity, we adopt two common methods: (i) alpha-based heterogeneity and (ii) sharding-based heterogeneity. In the alpha-based approach, most of the client's data comes from a single class, whereas in the sharding-based method, each client exclusively receives data from a specific subset of multiple classes. Other details are available in~\cite{p4_full}.}

\arxiv{To generate non-IID tasks and analyze the impact of data heterogeneity, we adopt two common methods: (i) alpha-based heterogeneity and (ii) sharding-based heterogeneity. Both partition data among clients to create non-IID distributions. In the alpha-based approach, most of the client's data comes from a single class, whereas in the sharding-based method, each client exclusively receives data from a specific subset of multiple classes. Sharding-based heterogeneity follows the procedure of~\cite{li2022learning}, where a dataset with $L$ classes is divided into $P$ shards per class, generating $M = \frac{LP}{N}$ tasks. Each task consists of $N$ randomly assigned classes, with one shard per class. We evaluate the effect of heterogeneity by setting $N \in \{2,4,8\}$. Alpha-based heterogeneity, used in prior work~\cite{noble2022differentially,hsu2019measuring}, controls heterogeneity through a parameter $\gamma$. For each client, $\gamma\%$ of the data is sampled IID from all classes, while the remaining $1-\gamma\%$ comes from a single dominant class. We assess heterogeneity effects using $\gamma \in \{25\%,50\%,75\%\}$.}

\subsubsection{Models}  
\label{subsec:hyperparams}

To assess the efficacy of our methods and compare them with related work, we conduct experiments using both a linear neural network with a softmax activation and a CNN-based architecture~\cite{tramer2020differentially}. Details about hyperparameters are in Appendix~\ref{apdx:hyperparams}\submit{ in the full paper~\cite{p4_full}}.

\subsubsection{Baselines}  

We select baselines based on the following criteria: (1) support for a peer-to-peer setting, (2) a privacy-preservation approach based on DP focused on privacy-utility amplification rather than an all-in-one system, and (3) publicly available code. Based on these criteria, we compare \name with the following baselines, which we adapt to our evaluation settings: \textit{centralized learning}, \textit{local training}, \textit{FedAvg}~\cite{mcmahan2017communication}, \textit{Scaffold}~\cite{noble2022differentially}, \textit{ProxyFL}~\cite{kalra2023decentralized}, \textit{DP-DSGT}~\cite{bayrooti2023differentially}. Appendix~\ref{apdx:exp_setup}\submit{ in the full paper~\cite{p4_full}} describes each baseline and how we adapt it to our evaluation settings, and provides additional experimental details.

\arxiv{
\begin{figure*}[t!]
    \centering
    \subfigure[]{\includegraphics[width=0.26\textwidth]{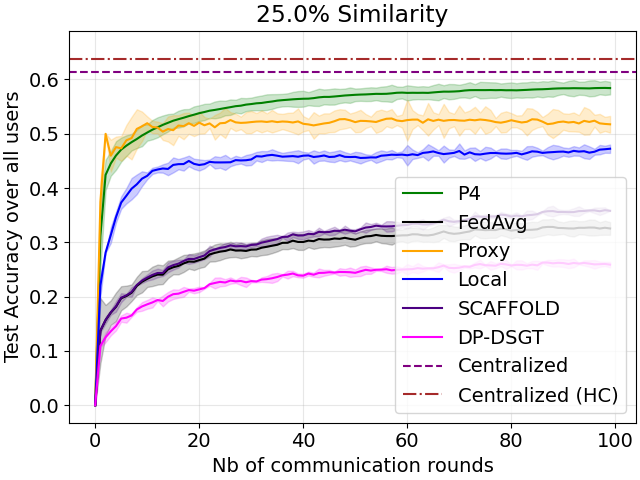}}
    \hspace{0.75cm}
    \subfigure[]{\includegraphics[width=0.26\textwidth]{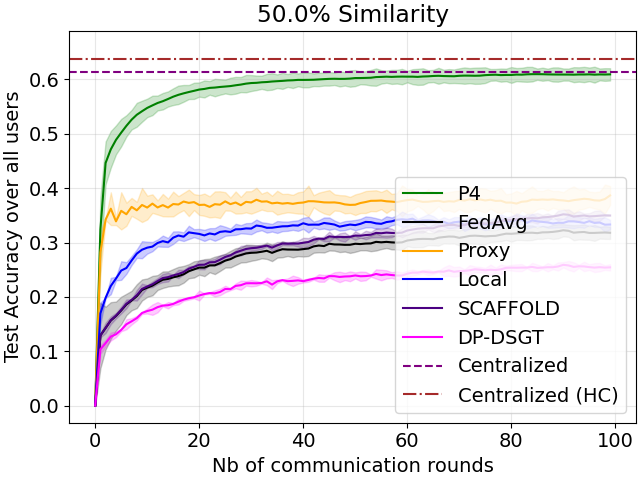}} 
    \hspace{0.75cm}
    \subfigure[]{\includegraphics[width=0.26\textwidth]{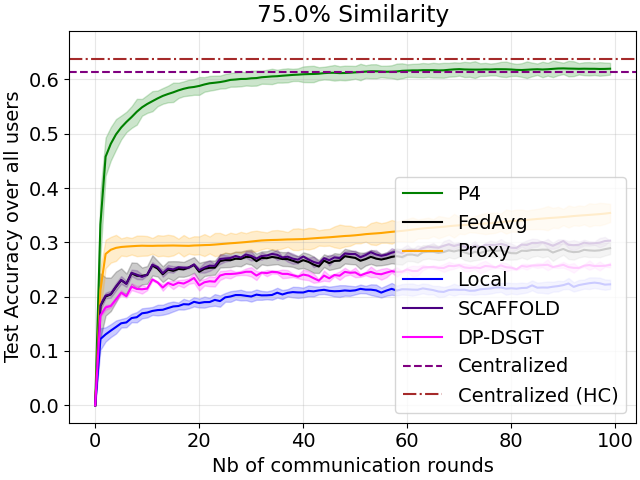}}
    \vspace{-0.2cm}
    \caption{Test accuracy of a linear model on \cifar with $\epsilon=15$ and alpha-based setting: (a) $\gamma=25\%$ (b) $\gamma=50\%$ (c) $\gamma=75\%$. For each client, $\gamma\%$ of the data is sampled IID from all classes, while the remaining $1-\gamma\%$ comes from a single dominant class.}
    \vspace{-0.25cm}
    \label{fig:diff_sim}
\end{figure*}

\begin{figure*}[t!]
    \centering
    \subfigure[]{\includegraphics[width=0.26\textwidth]{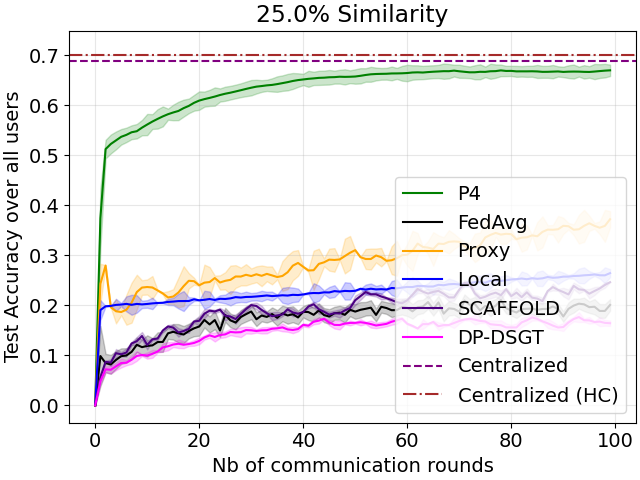}}
    \hspace{0.75cm}
    \subfigure[]{\includegraphics[width=0.26\textwidth]{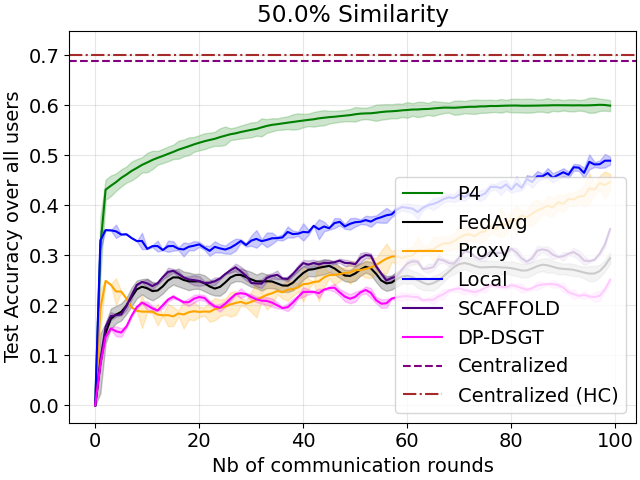}}
    \hspace{0.75cm}
    \subfigure[]{\includegraphics[width=0.26\textwidth]{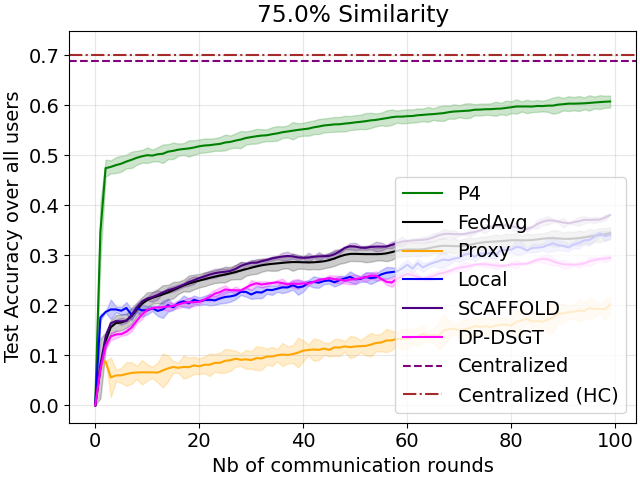}}
    \vspace{-0.2cm}
    \caption{Test accuracy of CNN on \cifar with $\epsilon=15$ and alpha-based non-IID setting: (a) $\gamma=25\%$ (b) $\gamma=50\%$ (c) $\gamma=75\%$. For each client, $\gamma\%$ of the data is sampled IID from all classes, while the remaining $1-\gamma\%$ comes from a single dominant class.}
    \vspace{-0.25cm}
    \label{fig:diff_sim_cnn}
\end{figure*}
}

\subsection{Privacy-Utility Amplification}
\label{sec:privacy_utility}

\submit{We first evaluate the performance of different algorithms under varying data heterogeneity by adjusting $\gamma$ and $N$, which correspond to alpha-based and shard-based heterogeneity, respectively. Performance under alpha-based heterogeneity is illustrated in Figures~\ref{fig:diff_sim}, 10 (linear model) and Figure~\ref{fig:diff_sim_cnn} (CNN model). Performance under shard-based heterogeneity is shown in Figures 8, 9 (Appendix~\ref{apdx:exp1:additional}) using a linear model on \cifarext and \femnist, respectively. In the figures, ``(HC)'' indicates the use of handcrafted features. Each experiment is repeated three times to mitigate randomness, and we report the mean results across trials. Other experimental details are available in the full paper~\cite{p4_full}.}

\arxiv{We first evaluate the performance of different algorithms under varying data heterogeneity by adjusting $\gamma$ and $N$, which correspond to alpha-based and shard-based heterogeneity, respectively. To simulate a scenario where clients have limited computational power, we fix the number of training rounds at $T = 100$, aiming for fast convergence on each client’s data distribution. For differential privacy, we set a target guarantee of $\epsilon=15$ for all methods. We consider all combinations of hyperparameters: local steps $K \in \{1,2,\dots,10\}$, client sampling ratio $l \in \{0.1,0.2,\dots,1\}$, and data sampling ratio $s \in \{0.1,0.2,\dots,1\}$, computing the corresponding noise level $\sigma_g$ using Equation~\ref{eq:dp_noise} to maintain the target $\epsilon$. Group sizes are set to $V=4$ for \cifarext and $V=8$ for other datasets. To ensure a fair comparison, we perform hyperparameter tuning (grid search) over the mentioned hyperparameters, along with local step size $\eta_l \in [0.1,10]$ and clipping norm $\mathcal{C} \in [0.1,10]$, using evaluation data from Section~\ref{sec:exp_setup}. The performance of different methods under shard-based heterogeneity is shown in Figure~\ref{fig:shard_FEMNIST_NN2} and Figure~\ref{fig:shard_CIFAR100_NN2} (Appendix~\ref{apdx:exp1:additional}) using a linear model on \femnist and \cifarext, respectively. Performance under alpha-based heterogeneity is illustrated in Figure~\ref{fig:diff_sim_cnn} (CNN model) and Figures~\ref{fig:diff_sim}, \ref{fig:emnist_results} (linear model, Appendix~\ref{apdx:exp1:additional}). In the figures, ``(HC)'' indicates the use of handcrafted features. Each experiment is repeated three times to mitigate randomness, and we report the mean results across trials.}

As shown in the figures, our approach consistently outperforms existing methods across different levels and types of data heterogeneity. On \cifar with alpha-based data generation, our method achieves $58.6\%$ to $62.2\%$ accuracy using only a single linear layer (see Figure~\ref{fig:diff_sim}), making it well-suited for resource-constrained IoT devices. In contrast, FedAvg struggles under data heterogeneity, underscoring the limitations of centralized federated learning with non-IID client data. Scaffold, another centralized method, performs better than FedAvg, particularly in highly heterogeneous settings (small $N$ and $\gamma$). However, despite its strengths in personalized federated learning, Scaffold still faces slow convergence, as observed across both data generation methods.

\arxiv{As shown in Figure~\ref{fig:shard_FEMNIST_NN2}, DP-DSGT, as a P2P method, achieves performance below the centralized baselines since its upper-bound performance is limited to achieving consensus among all clients (output of centralized methods). By considering the performance of FedAvg, Scaffold, and DP-DSGT on different datasets and architectures, we can conclude that when the tasks of clients are non-IID, consensus learning may not achieve good performance.}

In our experiments with a CNN model in Figure~\ref{fig:diff_sim_cnn}, which possesses more parameters compared to the linear model, previous approaches suffered from noisy training due to differential privacy noise and collaborative training with clients with different data distributions. In contrast, \name consistently converges to robust parameter solutions across all settings, even with a minimal number of communication rounds.

Our proposed model maintains stable performance across varying levels of data heterogeneity. Unlike other methods that perform well only under specific conditions---e.g., in Figure~\ref{fig:diff_sim_cnn}, Proxy~achieves good results at $\gamma=25\%$ but offers little improvement over local training in other cases---our approach consistently outperforms alternatives across both architectures and heterogeneity levels. A central factor contributing to this consistent performance is \name's client clustering strategy (see Section~\ref{subsection:similar_clients}).\submit{ Refer to Appendix~\ref{apdx:exp3:privacy} in the full paper~\cite{p4_full} for the ablation study on the privacy-utility improvement.}

\arxiv{Figure~\ref{fig:diff_sim} and Figures~\ref{fig:shard_CIFAR100_NN2}, \ref{fig:shard_FEMNIST_NN2}, \ref{fig:emnist_results} in Appendix~\ref{apdx:exp1:additional} show that when client's local dataset contains only a few classes and uses a simple neural network, \name can match or even surpass centralized training in some cases. This is because shallow networks more effectively learn patterns within a limited data distribution compared to the complexity of classifying all classes in a centralized setting. With fewer model parameters, classifying an entire data distribution becomes challenging for centralized methods, whereas a personalized model tailored to each client’s distribution shows even better performance.}

\arxiv{Our results suggest that local training can serve as a strong baseline for P2P learning when client data similarity is low. Thus, improving collaborative training in highly heterogeneous data distributions is crucial for achieving higher accuracy than local training. Appendix~\ref{apdx:exp1:comparison_local} provides an extended comparison of \name with local training under varying privacy budgets. In general, \name outperforms local training in accuracy, even when the privacy budget~exceeds~3.}

\submit{\subsection{Tolerating Poisoning Attacks}
\label{sec:tolerating_attacks}

In this section, we evaluate \name under the poisoning attacks described in Section~\ref{sec:threat_model} and client data heterogeneity. Our goal is to show that the methodology we use to improve the privacy-utility trade-off also enhances robustness against attacks. Additionally, we aim to show the effectiveness of the defense mechanism proposed in Section~\ref{subsection:malicious_security}. The full version~\cite{p4_full} provides a detailed analysis of how each component of our defense contributes to robustness, and demonstrates that our proposed defense is more effective than other defenses.

To evaluate \name’s robustness under different threat levels, we vary the proportion of malicious clients from 10\% to 50\%, reflecting typical ranges in privacy-preserving P2P systems from Section~\ref{sec:related_work}, and measure the impact of attacks on \name's performance. The experiment focuses on a linear model and \cifarext and \femnist datasets with $N = 4$ non-IID setting. Each experimental pipeline was run with three different seeds. As a performance metric, we measured the difference between the test accuracy of \name with an ``ideal defense'' and \name with the proposed secure aggregation. Here, the ``ideal defense'' assumes perfect knowledge of malicious clients and excludes them from aggregation. This metric is more suitable than \textit{attack impact}~\cite{xu2022byzantine} since the cumulative training and test sets for benign clients vary across different malicious client proportions, which may cause an unfair comparison. Thus, for each malicious percentage, we compare the ideal defense against \name's proposed~defense.

Figure~\ref{fig:ideal_vs_p4_defense} presents the results averaged across multiple seeds. Following the used benchmark~\cite{han2024fedsecurity}, an accuracy drop below 10\% indicates good defense performance. Based on this threshold and results across both datasets, \name with the proposed defense tolerates up to 50\% malicious clients under \lf, 30\% under \byz, 40\% under \byzr, and 30\% under \byzf. The likely reason for the high attack impact above 30\% of malicious clients is that \mk performs well only when the number of malicious clients is below $(n  - 2) / 2$, according to~\cite{blanchard2017machine}. See Appendix~\ref{apdx:exp3:robustness} in the full paper~\cite{p4_full} for an ablation study of \name's robustness and overhead when scaling to larger groups.

\begin{figure}[t!]
    \centering
    \includegraphics[width=\linewidth]{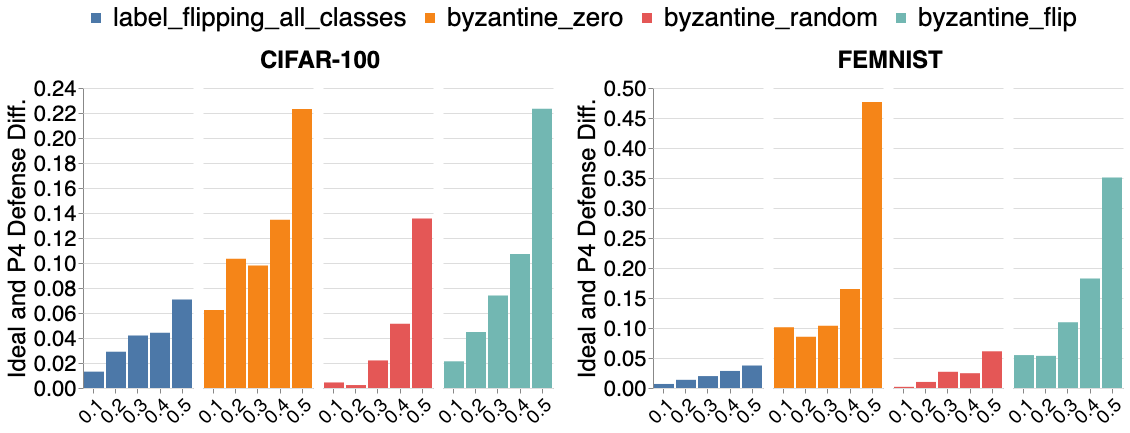}
    \vspace{-0.5cm}
    \caption{Differences of the test accuracy of an ideal defense and \name with secure aggregation on \cifarext, \femnist, and a linear model under malicious client percentages from 10\% to 50\%. Lower values indicate better attack tolerance.}
    \label{fig:ideal_vs_p4_defense}
    \vspace{-0.6cm}
\end{figure}
}\arxiv{\subsection{Tolerating Poisoning Attacks}
\label{sec:tolerating_attacks}

In this section, we evaluate \name under the poisoning attacks described in Section~\ref{sec:threat_model} and client data heterogeneity. Our goal is to demonstrate that the methodology we use to improve the privacy-utility trade-off also enhances robustness against attacks. Additionally, we aim to show the effectiveness of the defense mechanism proposed in Section~\ref{subsection:malicious_security}. To this end, we address the following research questions: (\texttt{RQ1}) How does client clustering in \name benefit robustness?; (\texttt{RQ2}) How does the combination of client clustering and proxy-based knowledge distillation enhance robustness compared to existing defense strategies?; (\texttt{RQ3}) What percentage of malicious clients can \name with a defense proposed in Section~\ref{subsection:malicious_security} tolerate?; (\texttt{RQ4}) Do \mk and \anom exhibit better robustness when used with \name compared to FedAvg?

Unless otherwise specified, all experiments use a linear model and the same datasets as previous evaluations, with one non-IID setting: 50\% similarity for \cifar and $N = 4$ for \cifarext and \femnist. The proportion of malicious clients is set to 30\% (as in~\cite{shayan2020biscotti, arapakis2023p4l}), and all clients participate in each iteration to simulate the strongest attack impact. Each experimental pipeline was run with three different seeds. For evaluation, we measure \textit{attack impact} (as in~\cite{xu2022byzantine}), defined as the test accuracy drop compared to the baseline (without any attack or defense), measured only on benign clients. This metric is chosen because the goal of these attacks is to degrade the quality of the final model.

We do not compare robustness of \name against P2P frameworks from Section~\ref{sec:related_work} that also consider differential privacy and malicious security for the following reasons: (1) these frameworks do not consider heterogeneous client data distributions in their methodology, therefore direct comparisons of robustness in terms of accuracy drop is unfair; (2) our goal is to show that the proposed privacy-utility amplification also enhances robustness, resulting in a good attack tolerance of \name, rather than to outperform other existing defenses; (3) our approach is orthogonal to existing DFL defenses, which also can be incorporated into \name instead of \mk; and (4) those privacy-preserving frameworks that address poisoning attack do not propose novel defense strategies, but rely on existing defenses. Instead, in this section we include these existing defenses in our evaluation, and in Section~\ref{sec:privacy_utility} we show that \name outperforms other state-of-the-art differential private P2P methods under various heterogeneity settings.

\paragraph{RQ1} We first evaluate \name without secure aggregation under poisoning attacks. The impact of model poisoning attacks on \name differs significantly before and after grouping (phase 1 in Section~\ref{sec:design}). Figure~\ref{fig:before_vs_after_grp} illustrates the attack impact on \name without secure aggregation, showing that after-grouping attacks have a much higher impact than before-grouping ones (similar to~\cite{li2021detection}). This occurs because, during group formation, \name detects malicious gradients based on $\ell_1$-norm similarity, clustering most malicious clients into separate groups that do not affect benign clients. We do not compare \lf before and after grouping, as we assume data poisoning always occurs before grouping. 
Since our threat model (Section~\ref{sec:threat_model}) assumes that an attacker may be aware of the group formation phase, we conduct further evaluations under the after-grouping setting for model poisoning attacks. To simulate this setting, clients are marked as malicious before group formation and start to behave maliciously during training (phase 2). As shown in Figure~\ref{fig:before_vs_after_grp} and dashed lines in Figure~\ref{fig:all_attacks}, \name without secure aggregation remains vulnerable to the selected model poisoning and data poisoning attacks. Among them, \lf has the smallest impact, while \byzr has the highest.

\begin{figure}[t!]
    \centering
    \includegraphics[width=0.85\linewidth]{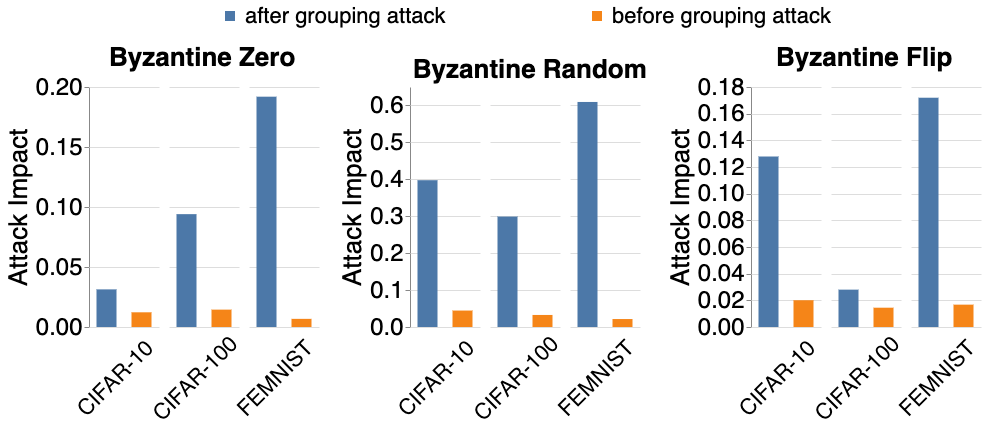}
    \vspace{-0.4cm}
    \caption{Attack impact on \name without secure aggregation before- and after-grouping for 3 non-IID datasets and a linear model with 30\% malicious clients.}
    \label{fig:before_vs_after_grp}
    \vspace{-0.5cm}
\end{figure}

\paragraph{RQ2} We compare the robustness of \name without secure aggregation (or simply client clustering with knowledge distillation) to nine existing defense strategies applied on top of \name. These defenses were selected based on their effectiveness in P2P settings and their compatibility with \name's privacy-utility logic (as explained in Section~\ref{subsection:malicious_security}). All defenses are implemented in the recent FedSecurity benchmark~\cite{han2024fedsecurity}, which we extend for our evaluation. Defense configurations follow those in the original papers~\cite{sun2019can, xie2019slsgd, yin2018byzantine, guerraoui2018hidden, pillutla2022robust, han2023kick, blanchard2017machine, fung2020limitations}. Figures~\ref{fig:all_attacks} and \ref{fig:all_attacks_big}(d) (results for \byzf are deferred to Appendix~\ref{apdx:exp2}) show attack tolerance of these defenses and \name without secure aggregation (\texttt{no\_defense}), where lower values indicate better performance. The x-axis orders the defenses based on the mean attack impact, and the dashed line shows the median of \texttt{no\_defense}.

\begin{figure}[t!]
    \centering
    \subfigure[\textbf{Label Flipping}]{\includegraphics[width=\linewidth]{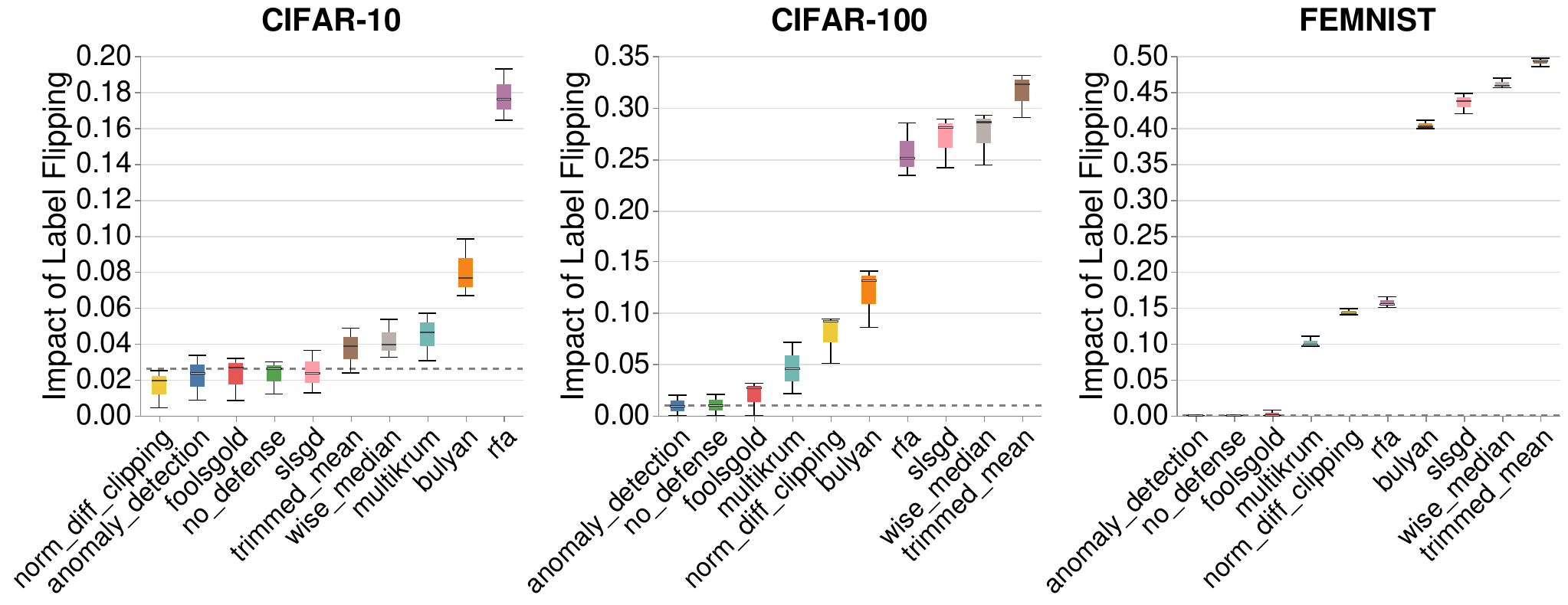}}
    \vspace{-0.1cm}
    
    \subfigure[\textbf{Byzantine Zero}]{\includegraphics[width=\linewidth]{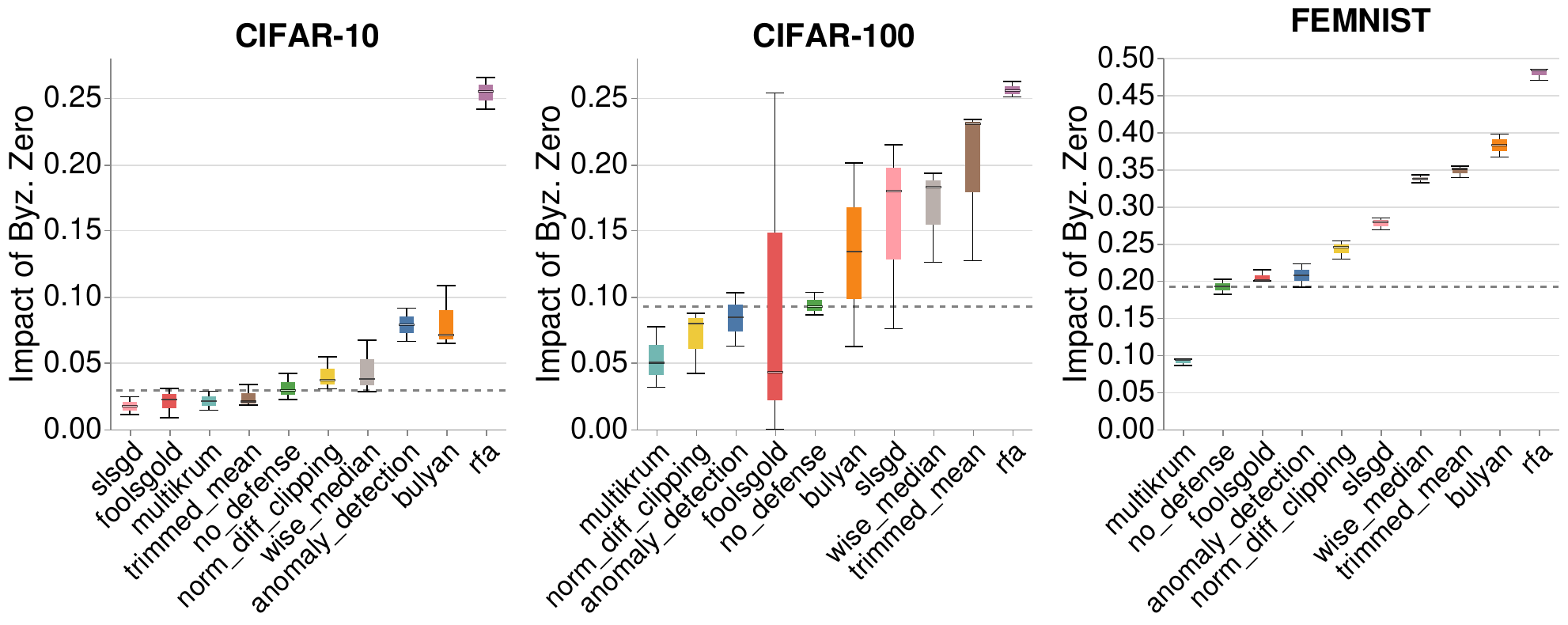}}
    \vspace{-0.1cm}
    
    \subfigure[\textbf{Byzantine Random}]{\includegraphics[width=\linewidth]{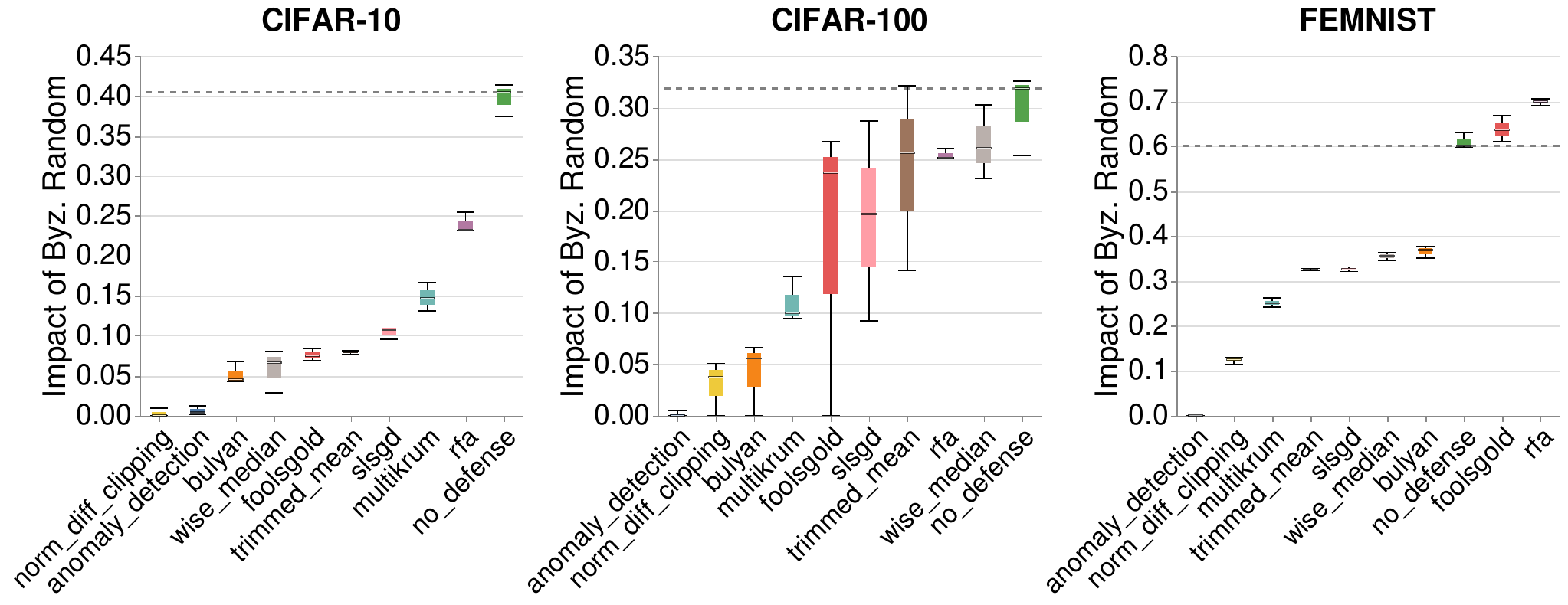}}

    \vspace{-0.25cm}
    \caption{Performance of \name with different defenses under poisoning attacks and 30\% of malicious clients.}
    \label{fig:all_attacks}
    \vspace{-0.55cm}
\end{figure}

A key observation is that \texttt{no\_defense} performs best under \lf and remains competitive against model poisoning attacks, supporting the effectiveness of \name’s client self-defense (mentioned in Section~\ref{subsection:malicious_security}). Notably, in Figure~\ref{fig:all_attacks}(a) \texttt{anomaly-detection} produces identical results to \texttt{no\_defense}, indicating its inability to detect anomalies in this setting and, thus, functioning as a normal \texttt{no\_defense}. Some defenses degrade \name’s accuracy compared to \texttt{no\_defense}, likely due to filtering benign clients (especially under \lf) or negatively impacting convergence. Figures~\ref{fig:all_attacks}(b) and \ref{fig:all_attacks_big}(d) show that \mk~\cite{blanchard2017machine} performs best against \byz and \byzf, with \texttt{Foolsgold}~\cite{fung2020limitations} being a close competitor, particularly in Figure~\ref{fig:all_attacks_big}(d). However, since \texttt{Foolsgold} performs worse on \femnist, \mk exhibits better overall attack tolerance under \byzf. The efficacy of \mk in combination with \name is discussed in Section~\ref{subsection:malicious_security}. For the most severe attack, \byzr, the simple and efficient \texttt{anomaly-detection}~\cite{han2023kick} achieves the best results. This may be because it uses model weight similarity during cross-client checks, resonating with \name’s clustering strategy. In most cases, \texttt{anomaly-detection} does not degrade \texttt{no\_defense} performance, making it a strong candidate for pairing with \mk.

\vspace{0.2cm}

\textit{RQ3.} To evaluate \name’s robustness under different threat levels, we combine its inherent client self-defense with \texttt{anomaly-detection} and \mk, using the former as a first-layer filter. We vary the proportion of malicious clients from 10\% to 50\%, reflecting typical ranges in privacy-preserving P2P systems from Section~\ref{sec:related_work}. The evaluation focuses on \cifarext and \femnist, which exhibited higher attack impact in RQ2. As a performance metric, we measured the difference between the test accuracy of \name with an ``ideal defense'' and \name with the proposed secure aggregation. Here, the ``ideal defense'' assumes perfect knowledge of malicious clients and excludes them from aggregation. This metric is more suitable than \textit{attack impact} since the cumulative training and test sets for benign clients vary across different malicious client proportions, which may cause an unfair comparison. Thus, for each malicious percentage, we compare the ideal defense against \name's proposed~defense.

Figure~\ref{fig:ideal_vs_p4_defense} presents the results averaged across multiple seeds. Following the used benchmark~\cite{han2024fedsecurity}, an accuracy drop below 10\% indicates good defense performance. Based on this threshold and results across both datasets, \name with the proposed secure aggregation tolerates up to 50\% malicious clients under \lf, 30\% under \byz, 40\% under \byzr, and 30\% under \byzf. The likely reason for the high attack impact above 30\% of malicious clients is that \mk performs well only when the number of malicious clients is below $(n  - 2) / 2$, according to~\cite{blanchard2017machine}.

Based on this evaluation, \name with secure aggregation effectively mitigates the impact of up to 30\% malicious clients across all four attack types and non-IID settings, while preserving privacy-utility amplification, which satisfies our threat model in Section~\ref{sec:threat_model}. Additionally, our evaluation includes a comparison with an ``ideal defense,'' which provides a more rigorous method for assessing attack tolerance under varying percentages of malicious clients. This approach is absent in other P2P frameworks from Section~\ref{sec:related_work} that provide a defense against poisoning attacks.

\begin{figure}[t!]
    \centering
    \includegraphics[width=\linewidth]{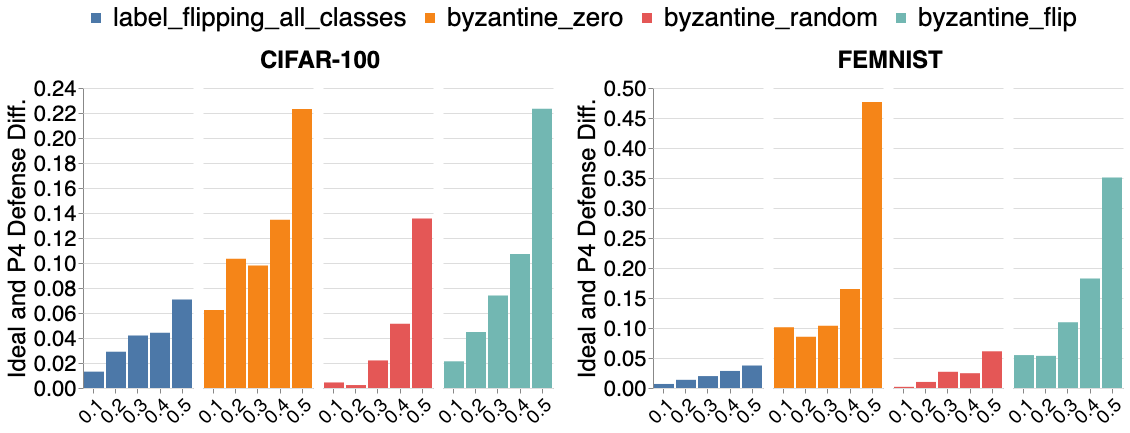}
    \vspace{-0.5cm}
    \caption{Differences of the test accuracy of an ideal defense and \name with secure aggregation on \cifarext, \femnist, and a linear model under malicious client percentages from 10\% to 50\%. Lower values indicate better attack tolerance.}
    \label{fig:ideal_vs_p4_defense}
    \vspace{-0.6cm}
\end{figure}

\vspace{0.2cm}

\textit{RQ4.} Due to space constraints, the comparison between \name and FedAvg under attacks is presented in Appendix~\ref{apdx:exp2:additional_results}. Overall, the results indicate that \name with a defense outperforms FedAvg with the same defense. Moreover, in most cases, the defense is more effective for \name than for FedAvg in terms of \textit{attack impact}, as it aligns with \name's client clustering.
}

\subsection{P4 on Resource-constrained Devices}
\label{subsection:energy}

We assess the practicality and overhead of \name by deploying it on resource-constrained devices. Specifically, we implement \name on two Raspberry Pi 4B clients running Debian 12 (Bookworm) 64-bit, Python 3.11.2, and PyTorch 2.1.0. The clients, using a linear model on \cifar, communicate via secure websockets. We evaluate four metrics: runtime, memory usage, power consumption, and communication overhead. Runtime is measured with Python's \texttt{time} library, memory usage with the \texttt{free} command, and power consumption using Tapo P110 smart plugs via the Tapo API. Communication overhead is assessed by measuring transmitted data between clients. The results for each metric are presented below.

\textit{Runtime.} We measure the average run times across 100 iterations each of both phases. We find that phase 1 (group formation) between two clients takes 0.04 seconds on average (std. 0.02). Assuming the same scenario as the experiment in Section~\ref{sec:privacy_utility} (each client samples 35 clients to compute its model similarity), it would take around 1.4 sec. in total to run phase 1. Phase 2 (co-training) between two clients takes an average of 5.27 sec. ($\pm$ 0.58) to complete, with the bulk of the runtime being the training process (avg. 4.83 s., std.~0.05). 

\textit{Memory usage.} We find that running \name consumes around 72 MB during phase 1 and 489 MB memory during phase 2.

\textit{Power consumption.} The baseline consumption of the Raspberry Pi is $\approx$ 2.64 W (std 0.01). We find that the average consumption during phase 1 is 3.17 W (std. 0.31) and phase 2 is 4.87 W (std. 0.34). 

\textit{Communication bandwidth.} In phase 1, the client that initiates the communication sends its model weights to another client, which then performs the comparison with its own weights. The message size of the weights is 622.82 kB (serialized with Python’s \texttt{pickle}). In phase 2, the client that initiates the co-training first sends its model parameters. After training, the recipient client sends back its gradients to the initiator client for aggregation. The total size of messages exchanged during this phase is 1246.57 kB. 

Our experiments indicate that \name can be effectively run on real-world IoT devices with minimal overhead.

\submit{\section{Conclusion}
\label{sec:conclusion}

This paper presents \name, a novel peer-to-peer learning framework tailored for IoT environments that simultaneously addresses the challenges of data heterogeneity, privacy preservation, and robustness against poisoning attacks. We propose a new lightweight, fully decentralized clustering algorithm to privately group similar clients, enabling efficient co-training via differentially private knowledge distillation. To defend against poisoning attacks, \name combines existing defenses such as \mk and \anom with client clustering and knowledge distillation. These components, originally designed to improve the privacy-utility trade-off, also enhance the effectiveness of the defense against malicious clients. Experiments demonstrate that \name outperforms state-of-the-art differentially private P2P methods by 5\%–30\% in accuracy and withstands up to 30\% malicious clients. Deployment on resource-constrained devices confirms its practicality for real-world IoT applications. Future work includes enabling secure, dynamic connectivity in decentralized settings by exploring blockchain-based trust mechanisms for client discovery, authentication, and communication integrity.
}\arxiv{\section{Conclusion}
\label{sec:conclusion}

This paper presents \name, a novel peer-to-peer learning framework that simultaneously addresses the challenges of data heterogeneity, privacy preservation, and robustness against poisoning attacks. We propose a new lightweight and fully decentralized client clustering algorithm that privately groups clients with similar model weights using the $\ell_1$-norm. Once grouped, clients engage in private co-training through differentially private knowledge distillation, enabling effective knowledge sharing while preserving individual data privacy. To defend against poisoning attacks, \name combines existing defenses such as \mk and \anom with client clustering and knowledge distillation. These components, originally designed to improve the privacy-utility trade-off, also enhance the effectiveness of the defense against malicious clients. Our evaluation shows that \name improves the privacy-utility trade-off, achieving 5\% to 30\% higher accuracy compared to state-of-the-art differentially private P2P methods, and tolerates up to 30\% malicious clients under various heterogeneity settings. Furthermore, our deployment on resource-constrained devices highlights \name's practicality and efficiency, making it a promising solution for real-world decentralized learning applications at the edge. Future work includes enabling secure, dynamic connectivity in decentralized settings by exploring blockchain-based trust mechanisms for client discovery, authentication, and communication integrity. We also plan to enhance resilience against poisoning attacks by advancing proxy-based knowledge distillation into a client-side self-defense mechanism and by mitigating threats from malicious aggregators through multi-aggregator verification and model consistency checks.
}

\bibliographystyle{IEEEtran}
\bibliography{base}

\clearpage

\appendices

\section{Table of Notations}
\label{apdx:notations}

Table~\ref{tab:notations} summarizes the main notations used throughout the paper.

\begin{table}[h!]
\centering
\caption{Summary of the main notations.}
\label{tab:notations}
\vspace{-0.25cm}

\begin{tabular}{ll}
\toprule
\textbf{Symbol} & \textbf{Description} \\
\midrule

$M,\ i \in [M]$ & number and index of clients \\

$M^{'}$ & number of gradients to be aggregated \\

$T,\ t \in [T]$ & number and index of communication rounds \\

$K,\ k \in [K]$ & number and index of local updates \\ & (for each client) \\

$D_i$ & local dataset held by the $i$-th client, composed \\ &of points $d_1^i, \ldots, d_R^i$ \\

$R$ & size of any local dataset $D_i$ \\

$D$ & joint dataset $\left(\bigsqcup_{i=1}^M D_i\right)$ \\

$\mathbf{C}$ & set of all possible class values in $D$ \\

$\theta^{(t)}$ & server model after round $t$ \\

$\theta_i^{(t)}$ & proxy model of $i$-th client after round $t$ \\

$\phi_i^{(t)}$ & private model of $i$-th client after round $t$ \\

$g_{ij}$ & gradient calculated on $i$-th client model on its \\ &$j$-th data sample \\

$G \in {0,1}^{M \times M}$ & binary matrix, where an entry $G_{ij}$ is one if \\&clients $i$ and $j$ are part of the same group, \\ &and zero otherwise \\

$V$ & group size \\

$l \in (0, 1)$ & client sampling ratio \\

$s \in (0, 1)$ & data sampling ratio \\

$\epsilon, \delta$ & differential privacy parameters \\

$\sigma_g$ & standard deviation of Gaussian noise added\\ &for privacy \\

$\mathcal{C} > 0$ & gradient clipping threshold \\

\bottomrule
\end{tabular}
\end{table}

\section{Additional Results for Privacy-Utility Amplification}
\label{apdx:exp1}

\subsubsection{Additional heterogeneity settings} 
\label{apdx:exp1:additional}

Figures~\ref{fig:shard_CIFAR100_NN2}, \ref{fig:diff_sim} and \ref{fig:emnist_results} illustrate the comparison of \name with existing privacy-preserving methods under additional heterogeneity settings for \cifarext, \cifar, and \femnist. The plots extend the results presented in Section~\ref{sec:privacy_utility}, and support the statement made in the main body that \name outperforms existing methods in terms of accuracy under different models, datasets, and data heterogeneity settings (alpha-based and shard-based).

\subsubsection{Comparison between collaborative and local training}
\label{apdx:exp1:comparison_local}

\begin{figure}[h!]
    \centering
    \includegraphics[width=\linewidth]{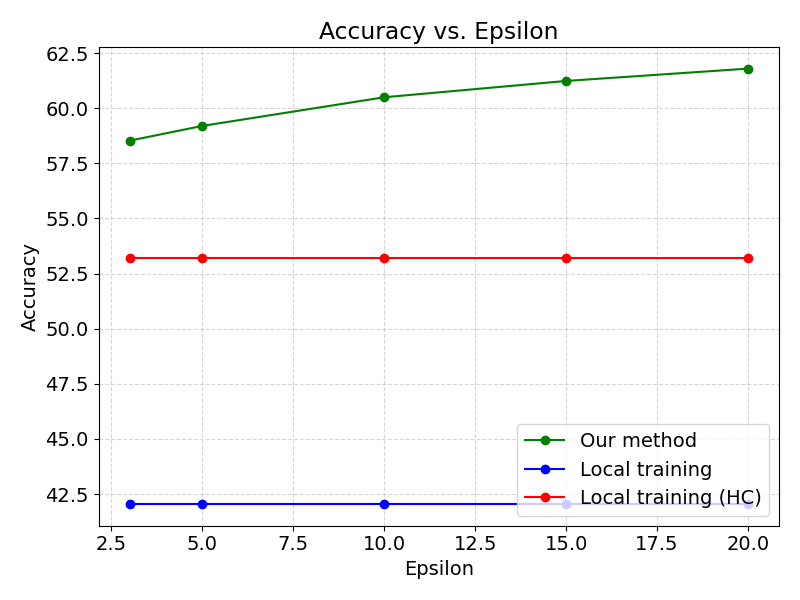}
    \vspace{-0.5cm}
    \caption{Performance of a linear model on \cifar under various privacy budgets and compared to local training.}
    \label{fig:tradeoff}
\end{figure}

\begin{figure*}[t!]
    \centering
    \subfigure[]{\includegraphics[width=0.30\textwidth]{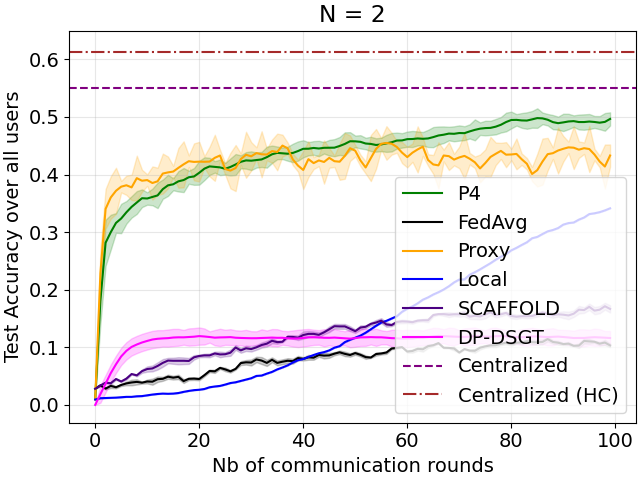}}
    \hspace{0.25cm}
    \subfigure[]{\includegraphics[width=0.30\textwidth]{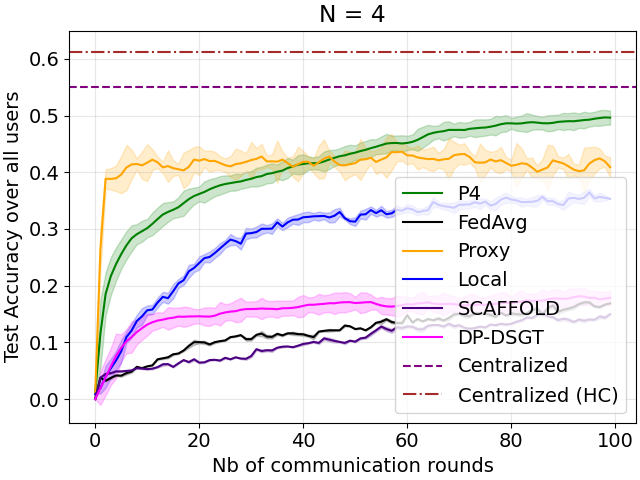}}
    \hspace{0.25cm}
    \subfigure[]{\includegraphics[width=0.30\textwidth]{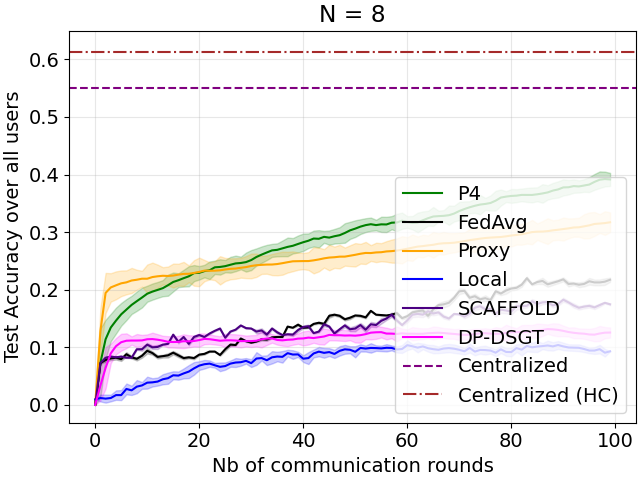}}
    \caption{Test accuracy of a linear model on \cifarext with $\epsilon=15$ and shard-based non-IID setting: (a) $N=2$ (b) $N=4$ (c) $N=8$.}
    \label{fig:shard_CIFAR100_NN2}
\end{figure*}

\begin{figure*}[t!]
    \centering
    \subfigure[]{\includegraphics[width=0.30\textwidth]{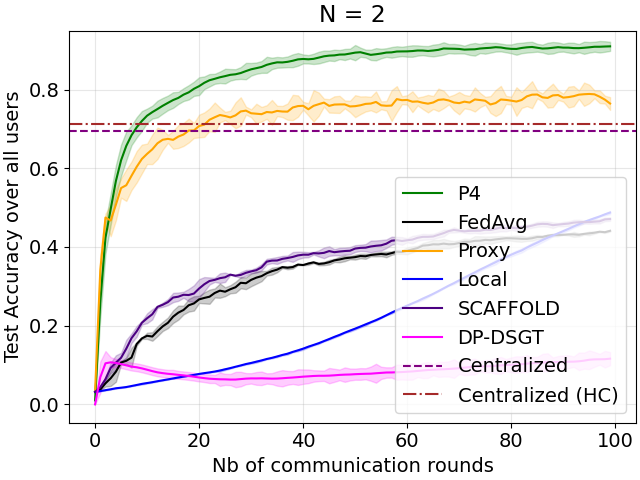}}
    \hspace{0.25cm}
    \subfigure[]{\includegraphics[width=0.30\textwidth]{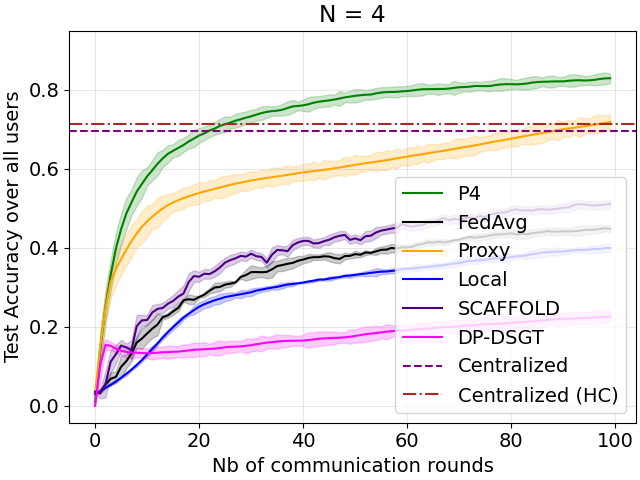}} 
    \hspace{0.25cm}
    \subfigure[]{\includegraphics[width=0.30\textwidth]{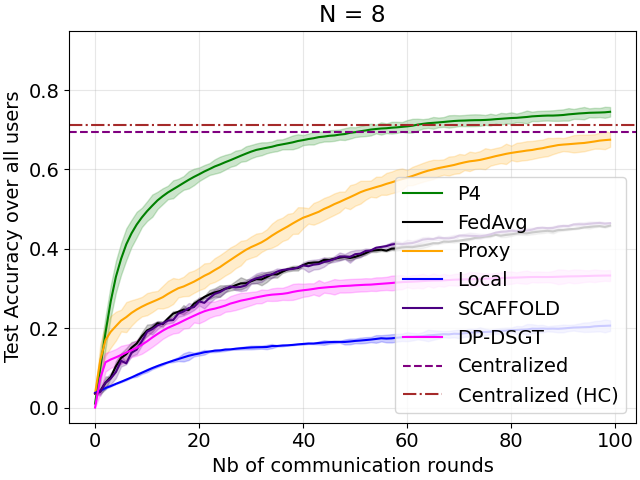}}
    \caption{Test accuracy of a linear model on \femnist with $\epsilon=15$ and shard-based non-IID setting: (a) $N=2$ (b) $N=4$ (c) $N=8$. Sharding-based heterogeneity divides a dataset with $L$ classes into $P$ shards per class, generating $M = \frac{LP}{N}$ tasks, where $M$ also equals the total number of clients. Each task consists of $N$ randomly assigned classes, with one shard per class.}
    \label{fig:shard_FEMNIST_NN2}
\end{figure*}

\begin{figure*}[t!]
    \centering
    \subfigure[]{\includegraphics[width=0.30\textwidth]{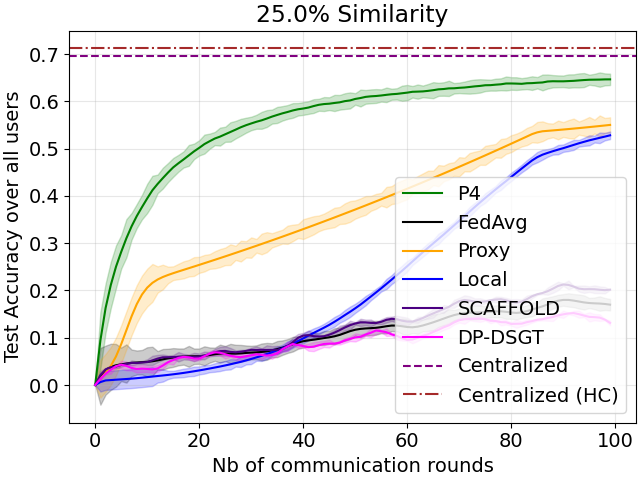}}
    \hspace{0.25cm}
    \subfigure[]{\includegraphics[width=0.30\textwidth]{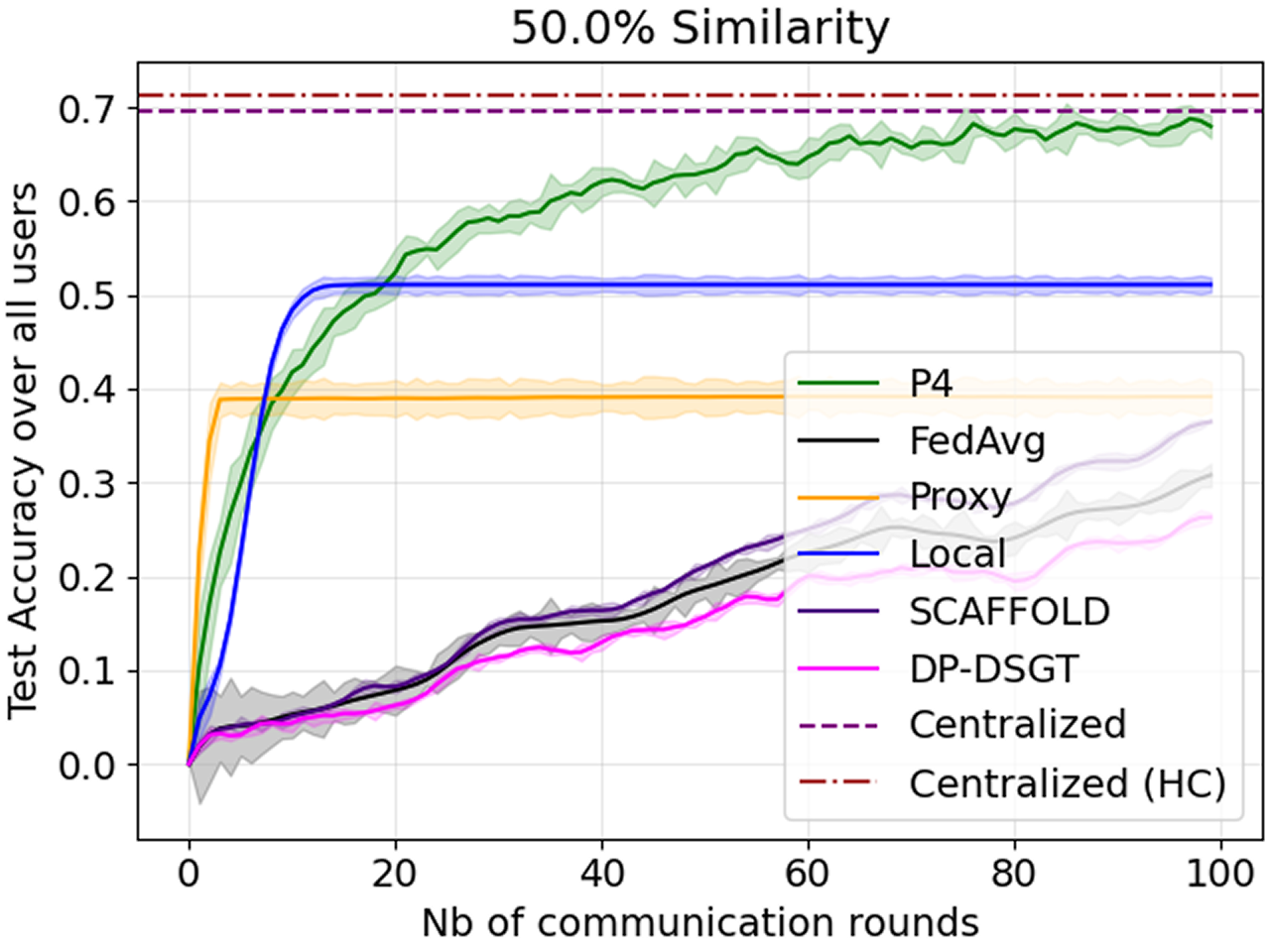}} 
    \hspace{0.25cm}
    \subfigure[]{\includegraphics[width=0.30\textwidth]{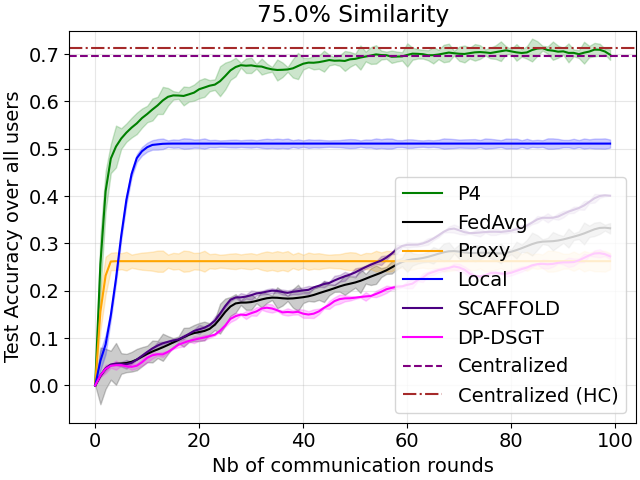}}
    \caption{Test accuracy of a linear model on \femnist with $\epsilon=15$ and alpha-based setting: (a) $\gamma=25\%$ (b) $\gamma=50\%$ (c) $\gamma=75\%$. For each client, $\gamma\%$ of the data is sampled IID from all classes, while the remaining $1-\gamma\%$ comes from a single dominant class.}
    \label{fig:emnist_results}
\end{figure*}

Distributed learning involves more communication and privacy concerns compared to local training. In local training, clients improve their models using only their own data, which means that they do not need to communicate with others during training. To see how well our differential private collaborative training method performs compared to the non-DP version of local training and to understand the impact on privacy, we tested our method under different privacy settings, with privacy bounds ranging from $\epsilon = 3$ to $\epsilon = 20$. The tasks are generated using the alpha-based method, and we utilized the FEMNIST dataset for this study. We compared this to how well local training performs on the previously mentioned linear model and data heterogeneity $\gamma = 50\%$. Our findings, illustrated in Figure~\ref{fig:tradeoff}, reveal that collaborative training outperforms local training even when the privacy budget is set higher than $3$. Even when strong privacy constraints are in place (e.g., $\epsilon = 3$), collaborative training performs better, addressing the issue of over-fitting that can affect local training due to its limited amount of local training data. Moreover, as shown in the figure, even though using handcrafted features could improve local training accuracy, it is still not as good as our proposed method. On the other hand, our method shows good robustness against DP noise such that it could have reasonable accuracy with restricted privacy.

\section{Additional Results for Tolerating Poisoning Attacks}
\label{apdx:exp2}

\subsubsection{Data poisoning tolerance}
\label{apdx:exp2:data_poisoning}

As shown in Figure~\ref{fig:ablation_defense_norm}, applying only a proxy mechanism is insufficient to fully mitigate data poisoning attacks. The figure illustrates the attack tolerance of different combinations of P4 components on the \cifar dataset when 30\% of clients are malicious. The attack impact percentage represents the difference in performance between the no-attack and attack scenarios, normalized by the no-attack performance to account for varying accuracy levels across different combinations. A lower attack impact percentage indicates better resilience.

The results demonstrate that merely incorporating a proxy mechanism does not guarantee robustness under arbitrary clustering settings (as indicated by the green bar). The primary reason that knowledge distillation fails under random clustering and non-IID conditions is analogous to why the classic FedAvg algorithm\arxiv{~\cite{mcmahan2017communication}} performs poorly in non-IID settings. Consider a scenario where a group consists of ten clients, with three being malicious and the remaining seven benign clients exhibiting distinct label distributions, \eg 90\% of local data corresponds to a single label, and this dominant label varies among the benign clients. As training progresses and malicious updates poison the proxy models, benign clients attempt to align their poisoned proxy models' softmax outputs with their \textit{local} models, guided by their local data distribution. However, when these partially cleansed proxy models are averaged, the differences in benign clients' data distributions diminish the overall effectiveness of the cleansing process. In contrast, when clients are clustered using \name's procedure, which leverages $\ell_1$-norm similarity, clients within the same group share a more similar data distribution. This grouping ensures that when proxy models --- cleaned based on their respective data distributions --- are aggregated, the effectiveness of data poisoning mitigation remains intact, improving attack resilience. Therefore, an appropriate group formation algorithm is vital for the aforementioned client self-defense.

\subsubsection{Additional results for Experiment 2}
\label{apdx:exp2:additional_results}

This subsection extends the results presented in Section~\ref{sec:tolerating_attacks}. Figure~\ref{fig:all_attacks_big}(d) illustrates the performance of \name with different defenses under a \byzf attack and 30\% of malicious clients (\textit{RQ2}). The plot supports the statement made in the main body that \mk shows better average performance than Foolsgold under \byzf, showing the similar attack impact on \cifar and \cifarext and on 8 points better tolerance on \femnist.

Figures~ \ref{fig:comparison_cifar_10} and \ref{fig:comparison_femnist} compare \name with a defense (proposed in Section~\ref{subsection:malicious_security}) against FedAvg with the same defense on \cifar and \femnist under different poisoning attacks (\textit{RQ4} in Section~\ref{sec:tolerating_attacks}). The results support the statement made in the main body that \name with a defense outperforms FedAvg with the same defense. Moreover, the proposed defense works better together with \name than with FedAvg, especially on \femnist, comparing the differences between ''\name \& no\_defense'' / ''\name \& defense'' and ''FedAvg \& no\_defense'' / ''FedAvg \& defense''. This highlights that the proposed defense aligns with \name's design and that personalization may also benefit attack tolerance.

\section{Ablation Studies}
\label{apdx:exp3}

\subsection{Privacy-Utility Improvement}
\label{apdx:exp3:privacy}

\begin{figure}[t!]
    \centering
    \includegraphics[width=\linewidth]{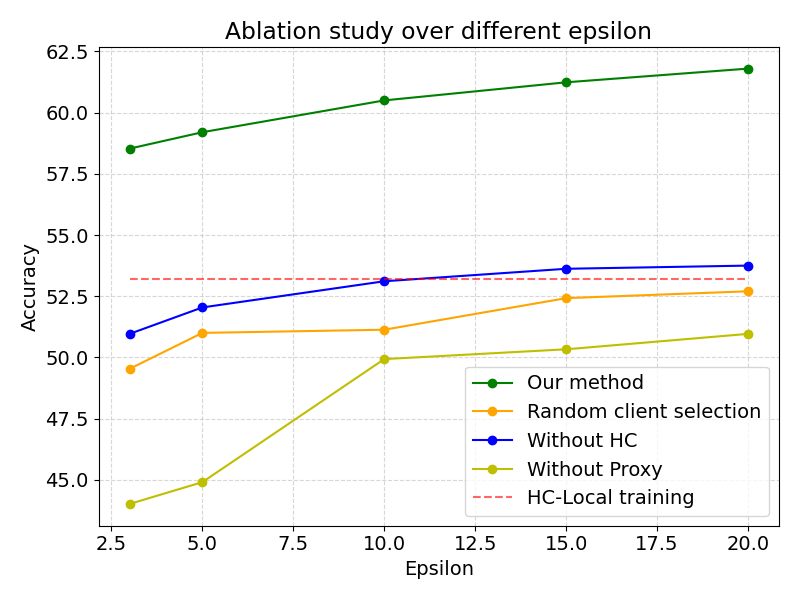}
    \vspace{-0.5cm}
    \caption{Comparing the effect of each component of \name on \cifar and a linear model under data similarity of $\gamma = 50\%$ and a privacy budget $\epsilon$ from 3 to 20.}
    \label{fig:ablation}
    \vspace{-0.6cm}
\end{figure}

\begin{figure}[t!]
    \centering
    \includegraphics[width=\linewidth]{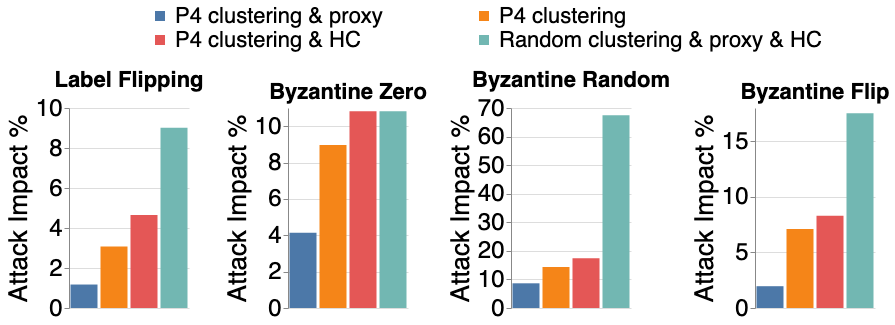}
    \vspace{-0.5cm}
    \caption{Impact of \name components on attack tolerance for \cifar with a linear model and 30\% malicious clients. Attack impact is normalized by test accuracy; lower values indicate better tolerance.}
    \label{fig:ablation_defense_norm}
    \vspace{-0.25cm}
\end{figure}

\begin{figure}[t!]
    \centering
    \includegraphics[width=\linewidth]{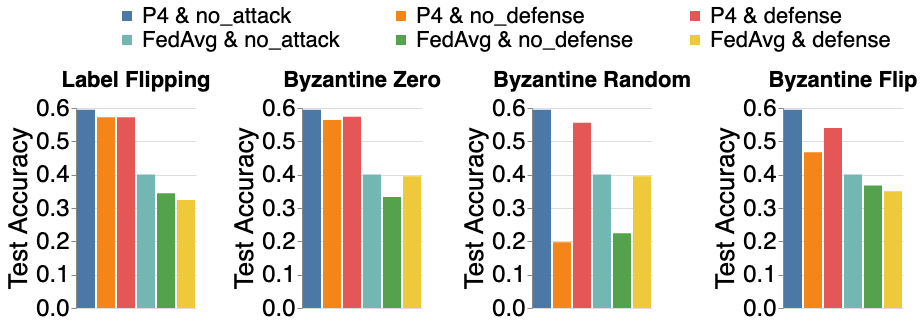}
    \vspace{-0.5cm}
    \caption{Comparison of \name with a defense against FedAvg for \cifar and a linear model under 30\% of malicious clients.}
    \label{fig:comparison_cifar_10}
\end{figure}

\begin{figure}[t!]
    \centering
    \includegraphics[width=\linewidth]{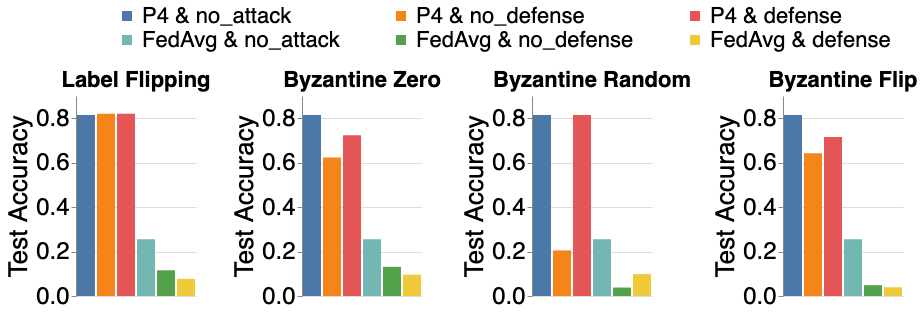}
    \vspace{-0.5cm}
    \caption{Comparison of \name with a defense against FedAvg for \femnist and a linear model under 30\% of malicious clients.}
    \label{fig:comparison_femnist}
\end{figure}

We now perform an ablation study of \name on \cifar with 50\% similarity using the linear model. We first aim to understand the effect of client selection, handcrafted features, and proxy model individually on the performance of \name. Therefore, we compare \name's accuracy results without secure aggregation with three different methods: i) random client selection instead of using our group clustering technique; ii) using raw images instead of handcrafted features; iii) removing the proxy model and using one model per client instead. In each experiment, one of these components is removed and the performance of the model is shown in Figure \ref{fig:ablation}. As shown in the figure, each of these three components has a strong effect on the model performance -- removing even one of them results in accuracy lower than the local training baseline in most cases. Our study shows that we can achieve private personalized learning only when all design components of \name are enabled.

\subsection{Robustness}
\label{apdx:exp3:robustness}

\begin{figure}[t!]
    \centering
    \subfigure[Varying the number of clients for group size = 8]{\includegraphics[width=0.46\linewidth]{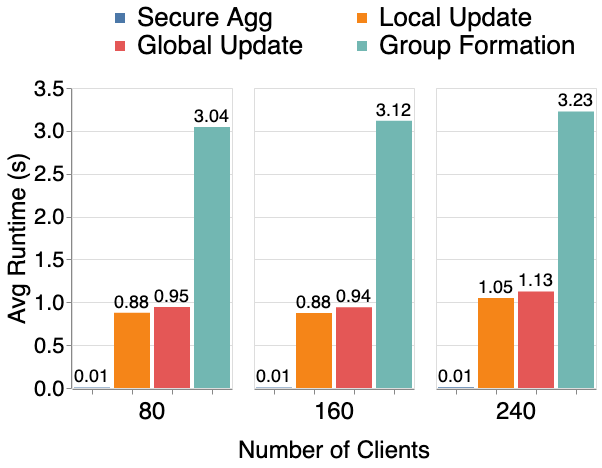}}
    \subfigure[Varying group size for 240 clients]{\includegraphics[width=0.46\linewidth]{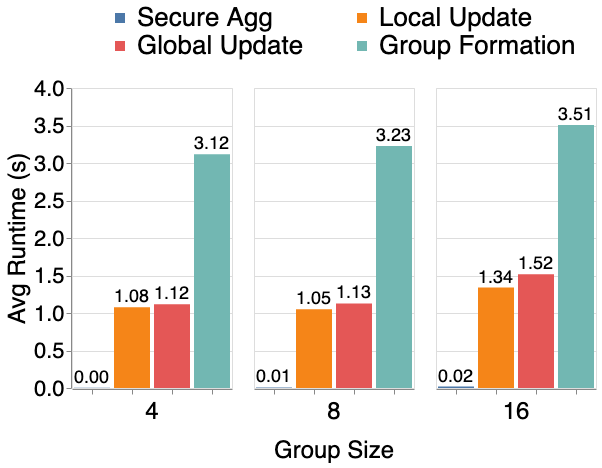}} 
    \caption{Breakdown of time spent in different mechanisms in \name on a linear model and \cifar with 50\% alpha-based similarity under 30\% of \byzr clients.}
    \vspace{-0.35cm}
    \label{fig:ablation_runtime}
\end{figure}

\begin{figure}[h]
    \centering
    \subfigure[]{\includegraphics[width=0.48\linewidth]{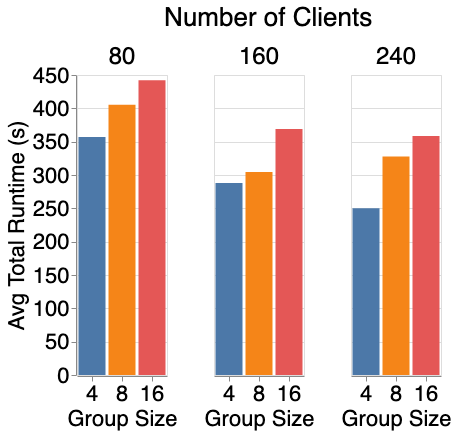}}
    \subfigure[]{\includegraphics[width=0.48\linewidth]{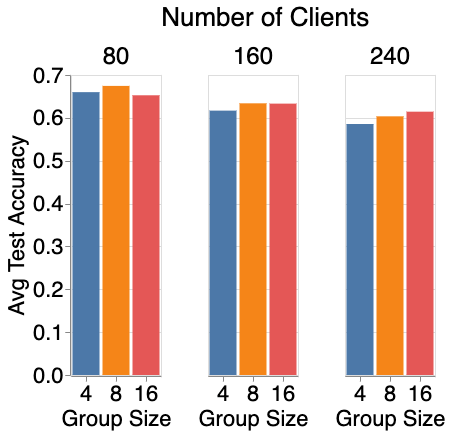}} 
    \caption{Average total runtime (in seconds) vs. test accuracy across different numbers of clients and group sizes for \cifar and a linear model. A dataset size is the same across different numbers of clients, indicating more samples per client for a smaller total number of clients.}
    \label{fig:total_runtime}
\end{figure}

\begin{figure*}[h!]
    \centering
    \subfigure[\textbf{Label Flipping}]{\includegraphics[width=0.7\linewidth]{figures/exp2/lf_all_defenses_v3.pdf}}
    \vspace{-0.1cm}
    
    \subfigure[\textbf{Byzantine Zero}]{\includegraphics[width=0.7\linewidth]{figures/exp2/byz_zero_all_defenses_v3.pdf}}
    \vspace{-0.1cm}
    
    \subfigure[\textbf{Byzantine Random}]{\includegraphics[width=0.7\linewidth]{figures/exp2/byz_rand_all_defenses_v3.pdf}}
    \vspace{-0.1cm}

    \subfigure[\textbf{Byzantine Flip}]{\includegraphics[width=0.7\linewidth]{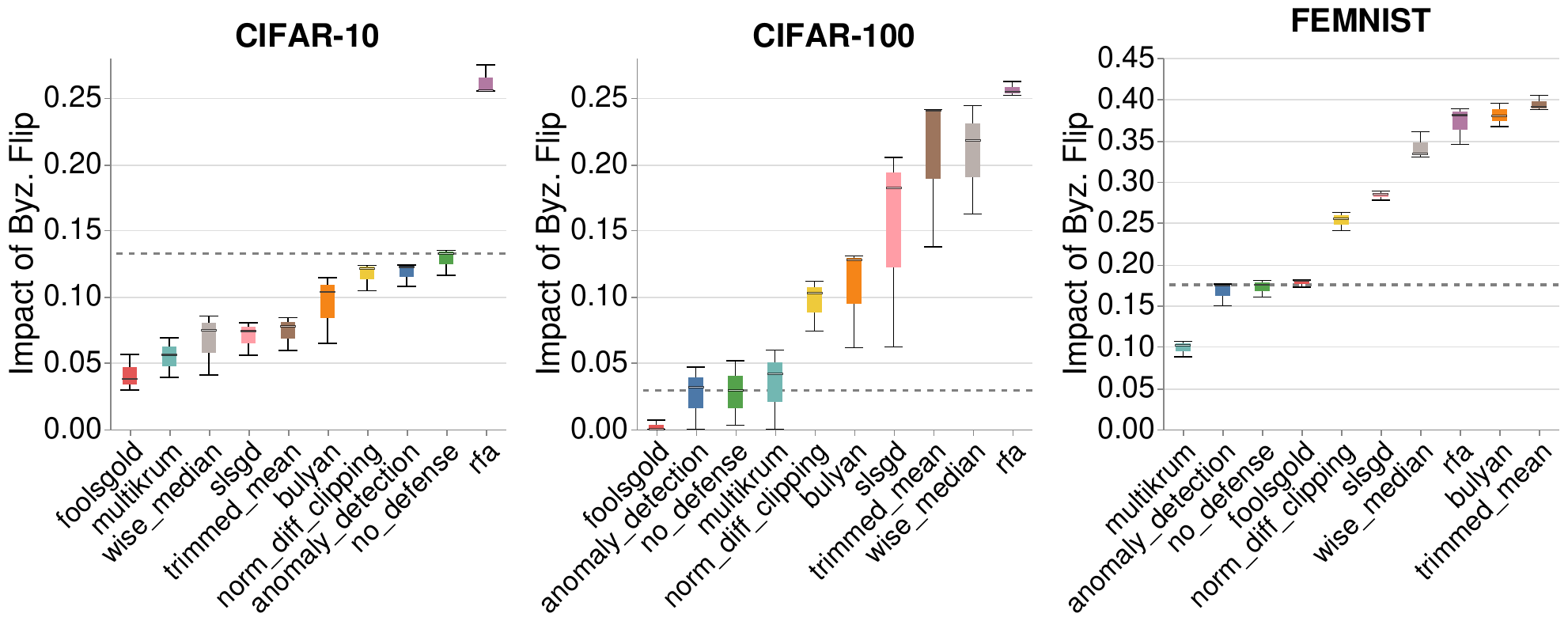}}

    \vspace{-0.25cm}
    \caption{Performance of \name with different defenses under all four poisoning attacks and 30\% of malicious clients.}
    \label{fig:all_attacks_big}
\end{figure*}

In this section, we provide an extended analysis of runtimes and test accuracy. We evaluate the impact of increasing the number of clients (Figure~\ref{fig:ablation_runtime}(a)), group size (Figure~\ref{fig:ablation_runtime}(b)), and the number of samples per client on \name (Figure~\ref{fig:total_runtime}). The experiment was conducted under a \byzr attack using the proposed defense from Section~\ref{subsection:malicious_security}, with a client participation ratio of 0.5 per global update and a fixed number of samples per client across all settings. Due to the large number of clients and configuration combinations, we performed this experiment in simulation, while the experiment in Section~\ref{subsection:energy} evaluates \name under real-world conditions. The runtime for group formation is measured in simulation without considering its distributed execution on multiple clients, therefore see the runtime for phase 1 in Section~\ref{subsection:energy} for realistic numbers.

Figure~\ref{fig:ablation_runtime} presents a breakdown of execution time (in seconds) across different \name mechanisms. From Figure~\ref{fig:ablation_runtime}(a), we observe that even when combining \texttt{anomaly-detection} and \mk, secure aggregation remains efficient and is influenced more by group size than by the total number of clients. This is because, unlike centralized aggregation, \name performs aggregation on multiple aggregators (one per group), where group sizes remain relatively small. Additionally, the global update per group introduces minimal overhead compared to an average local update, with its runtime primarily dependent on group size. Since this experiment assumes distributed training across clients, the runtime per global update comprises the local update time for a single client, secure aggregation time, and minor overhead for group management on the aggregator. With multiple aggregators and small group sizes, the runtime difference between a global and local update remains small. Group formation runtime is also more dependent on group size than the total number of clients, though its variation across different settings in Figure~\ref{fig:ablation_runtime} remains minimal. The observed inconsistency in local update runtime across different group sizes likely stems from how our code computes the average runtime per client within a group before averaging across all groups. Larger group sizes lead to greater variance in local update runtimes. Figure~\ref{fig:total_runtime} further illustrates total runtime and test accuracy under varying group sizes and client numbers. Since for this plot the total dataset size remains constant, a smaller number of clients results in more samples per client. The results indicate a trade-off between total runtime and test accuracy, with larger group sizes increasing runtime. Thus, selecting an appropriate group size requires careful consideration of system constraints.

\vspace{2.0cm}

\section{Additional Experimental Setup}
\label{apdx:exp_setup}

\subsubsection{Models and Hyperparameters}  
\label{apdx:hyperparams}

To assess the effectiveness of our methods and compare them with related work, we conduct experiments using both a shallow neural network (a linear layer with a softmax activation) and a CNN-based architecture\arxiv{~\cite{tramer2020differentially}}. To ensure fair comparisons, we perform parameter tuning within a fixed range across all methods. Following\arxiv{~\cite{noble2022differentially}}, we set the global step size to $\eta_g=1$ and define the local step size as $\eta_l=\frac{\eta_0}{sK}$, where $\eta_0$ is carefully tuned. For privacy, we maintain a fixed $\delta=\frac{1}{R}$ in all experiments. Given the fixed number of global training rounds $T$ and a target privacy budget $\epsilon$, we tune $K$, $l$, $s$ values, computing the corresponding noise level $\sigma_g$ using Equation~\ref{eq:dp_noise}.

\subsubsection{Baselines}  

We select baselines based on the following criteria: (1) support for a peer-to-peer setting, (2) a privacy-preservation approach focused on privacy-utility amplification rather than an all-in-one system, and (3) publicly available code. Based on these criteria, we compare \name with the following baselines:

\vspace{0.10cm}

\noindent \textbf{Centralized learning:} In this setting, all data is stored in a central location without privacy concerns, and a high-computational server trains the model on the entire dataset. This serves as an optimal benchmark, reflecting the classification difficulty. We report results with and without handcrafted features.  

\vspace{0.10cm}

\noindent \textbf{Local training:} Each client trains a model solely on its local dataset. While this may seem like a weaker baseline, our results show that under high data heterogeneity, knowledge sharing provides only marginal improvements. In such cases, many prior approaches fail to surpass local training in accuracy.  

\vspace{0.10cm}

\noindent \textbf{FedAvg\arxiv{~\cite{mcmahan2017communication}}:} A widely-used federated learning algorithm where a centralized server coordinates training. The server is assumed to be honest-but-curious, so we compute the corresponding privacy bound $\epsilon$ with respect to the server.  

\vspace{0.10cm}

\noindent \textbf{Scaffold\arxiv{~\cite{noble2022differentially}}:} A centralized federated approach designed to handle data heterogeneity under DP constraints. By introducing an additional control parameter per client, Scaffold mitigates client drift from the global model, improving personalization over FedAvg. We assume that the server is untrustworthy, and privacy loss is computed using the sub-sampled Gaussian mechanism under Renyi Differential Privacy (RDP)\arxiv{~\cite{mironov2017renyi}}.  

\vspace{0.10cm}

\noindent \textbf{ProxyFL\arxiv{~\cite{kalra2023decentralized}}:} A state-of-the-art decentralized learning method where clients maintain a \textit{private} model and share a \textit{proxy} model with others. We assume the private and proxy models share the same architecture and follow the directed exponential graph communication method and differential privacy mechanism described in the ProxyFL paper.  

\vspace{0.10cm}

\noindent \textbf{DP-DSGT\arxiv{~\cite{bayrooti2023differentially}}:} One of the state-of-the-art approaches in differentially private P2P learning. Among the three methods proposed in\arxiv{~\cite{bayrooti2023differentially}}, DP-DSGT is designed for consistent performance across varying data distributions, making it well-suited for scenarios where clients have non-overlapping data classes.

Although previous studies have explored client clustering in decentralized learning\submit{~\cite{li2022towards}}\arxiv{~\cite{li2022towards,duan2021flexible,xie2021multi,sattler2020clustered,nguyen2022self,briggs2020federated,ghosh2020efficient}}, these methods either rely on a centralized server for clustering or violate differential privacy. Thus, we exclude them from our comparison with \name. We also do not compare with AvgPush\footnote{A decentralized version of FedAvg using PushSum for aggregation.}, CWT\arxiv{~\cite{chang2018distributed}}, and FML\arxiv{~\cite{shen2020federated}} since ProxyFL has already been shown to outperform them in\arxiv{~\cite{kalra2023decentralized}}. Additionally, we exclude SMC-based techniques\arxiv{~\cite{kasyap2021privacy, truex2019hybrid}}, as these methods are often computationally intensive due to their reliance on cryptographic primitives such fully homomorphic encryption.

\subsubsection{Parameter Settings}  

As described in Section~\ref{subsection:private_train}, in some experiments we use handcrafted features instead of raw images. Specifically, following\arxiv{~\cite{tramer2020differentially}}, we adopt the scattering network of\arxiv{~\cite{oyallon2015deep}}, which encodes images using wavelet transforms\arxiv{~\cite{bruna2013invariant}}. Using the default parameters from\arxiv{~\cite{oyallon2015deep}}, we employ a ScatterNet $S(x)$ of depth two with wavelets rotated along eight angles. For an image of size $H \times W$, the handcrafted feature extractor outputs $(K,\frac{H}{4},\frac{W}{4})$, where $K=81$ for grayscale images and $K=243$ for RGB images. Additionally, we apply data normalization, where each client computes the mean and variance of its local data, incurring no additional privacy cost. Using these transformed features enhances accuracy and allows classification with a single linear layer, as demonstrated in Section~\ref{sec:privacy_utility}.

\section{Why One Iteration of Local Training is Sufficient to Distinguish Different Distributions}
\label{sec:one_epoch_distinguish}

\paragraph{Goal.} We show that if two clients have sufficiently different data distributions, then even a single iteration of local training (under differential privacy) produces updated model weights that are distinguishable with high probability. In particular, when the difference in their expected gradients dominates the sampling, noise, and clipping errors, the $\ell_1$ distance between their weight vectors (cf. Equation~\eqref{eq:l1_similarity_restated}) remains large. This observation underpins the clustering strategy described in Section~\ref{subsection:similar_clients}.

\subsection{Setup and Notation}

Let $\mathcal{D}_i$ and $\mathcal{D}_j$ denote the data distributions for clients $i$ and $j$, respectively. Each client draws its local dataset i.i.d.\ from its distribution. All clients start from a common initial weight vector $w_0$ (or $\theta_0$) and perform one iteration of gradient-based training with differential privacy (DP). After one iteration, the updated weights for clients $i$ and $j$ are denoted by $w_i$ and $w_j$, respectively.

The dissimilarity metric is given by
\begin{equation}
\label{eq:l1_similarity_restated}
\mathrm{dissimilarity}(i,j) = \Bigl\| \mathbf{w}_i - \mathbf{w}_j \Bigr\|_1.
\end{equation}

For each client, let
\[
g_i = \frac{1}{n_i} \sum_{t=1}^{n_i} \nabla \ell(w_0; x_t, y_t)
\]
be the iteration-averaged empirical gradient (with $n_i$ samples) and define its expectation as
\[
\mathbb{E}[g_i] = \mathbb{E}_{(x,y) \sim \mathcal{D}_i} \Bigl[\nabla \ell(w_0; x,y)\Bigr].
\]
We assume that the expected gradients differ by at least
\[
\|\mathbb{E}[g_i] - \mathbb{E}[g_j]\|_1 \ge \gamma,
\]
for some $\gamma>0$, which reflects the intrinsic difference in the underlying data distributions.

After gradient clipping with threshold $\mathcal{C}$ and adding Gaussian noise for DP (with noise standard deviation $\sigma = \tilde{\sigma}\,\mathcal{C}$), the communicated gradient for client $i$ is
\[
\widetilde{g}_i = \mathrm{clip}(g_i, \mathcal{C}) + \xi_i,
\]
where $\xi_i \sim \mathcal{N}(0,\sigma^2 I_d)$. The corresponding weight update is given by
\[
w_i = w_0 - \eta\, \widetilde{g}_i,
\]
with learning rate $\eta$.

\subsection{Assumptions}

We make the following assumptions:

\begin{enumerate}[label=(A\arabic*)]
    \item \textbf{Gradient Clipping and Clipping Error.} For each client, define the clipping error as 
    \[
    \epsilon_{\mathrm{clip}} = \mathrm{clip}(g,\mathcal{C}) - g,
    \]
    where
    \[
    \mathrm{clip}(g,\mathcal{C}) = \min\Bigl(1, \frac{\mathcal{C}}{\|g\|_2}\Bigr) g.
    \]
    We do not assume that clipping is always inactive; instead, standard tail bounds (e.g., under a sub-Gaussian assumption on $g$) ensure that
    \[
    \mathbb{E}\bigl[\|\epsilon_{\mathrm{clip}}\|_1\bigr] \le \mathbb{E}\Bigl[\|g\|_1\,\mathbf{1}_{\{\|g\|_2 > \mathcal{C}\}}\Bigr],
    \]
    so that for appropriate choices of $\mathcal{C}$ the clipping error is small.
    
    \item \textbf{Iteration-Averaged Gradients.} For each client $i$, the empirical gradient is defined as
    \[
    g_i = \frac{1}{n_i}\sum_{t=1}^{n_i}\nabla \ell(w_0;x_t,y_t),
    \]
    with expectation $\mathbb{E}[g_i]$ as above.
    
    \item \textbf{Concentration for Sampling and Noise.} There exist constants $c_1$ and $c_2$ such that, with probability at least $1-\delta$, 
    \begin{align*}
    \|g_i - \mathbb{E}[g_i]\|_1 &\le \frac{c_1}{\sqrt{n_i}}, \quad \|g_j - \mathbb{E}[g_j]\|_1 \le \frac{c_1}{\sqrt{n_j}},\\[1mm]
    \|\xi_i - \xi_j\|_1 &\le c_2\,\tilde{\sigma}\,\mathcal{C}\,\sqrt{d}.
    \end{align*}
    
    \item \textbf{Intrinsic Gradient Difference.} The expected gradients satisfy
    \[
    \|\mathbb{E}[g_i]-\mathbb{E}[g_j]\|_1 \ge \gamma,
    \]
    for some $\gamma>0$.
\end{enumerate}

\subsection{Proof}

\textbf{Step 1. Expressing the Weight Difference.} \\[1mm]
Each client updates its weight vector as
\[
w_i = w_0 - \eta\,\widetilde{g}_i, \quad w_j = w_0 - \eta\,\widetilde{g}_j,
\]
where the noisy gradient is given by
\[
\widetilde{g}_i = \mathrm{clip}(g_i,\mathcal{C}) + \xi_i = g_i + \epsilon_{\mathrm{clip}}^{(i)} + \xi_i.
\]
Thus, the difference in weights is
\[
w_i - w_j = -\eta \Bigl[(g_i-g_j) + \bigl(\epsilon_{\mathrm{clip}}^{(i)}-\epsilon_{\mathrm{clip}}^{(j)}\bigr) + (\xi_i-\xi_j)\Bigr].
\]
Taking the $\ell_1$ norm and using homogeneity yields
\begin{equation}
\label{eq:weight_diff_final}
\|w_i-w_j\|_1 = \eta\,\|\widetilde{g}_i-\widetilde{g}_j\|_1.
\end{equation}

\textbf{Step 2. Lower-Bounding $\|\widetilde{g}_i-\widetilde{g}_j\|_1$.} \\[1mm]
By the triangle inequality,
\begin{equation}
\label{eq:triangle_ineq}
\|\widetilde{g}_i-\widetilde{g}_j\|_1 \ge \|g_i-g_j\|_1 - \|\epsilon_{\mathrm{clip}}^{(i)}-\epsilon_{\mathrm{clip}}^{(j)}\|_1 - \|\xi_i-\xi_j\|_1.
\end{equation}
We now decompose the empirical gradient difference as
\[
g_i-g_j = \Bigl[\mathbb{E}[g_i]-\mathbb{E}[g_j]\Bigr] + \Bigl[(g_i-\mathbb{E}[g_i]) - (g_j-\mathbb{E}[g_j])\Bigr].
\]
Applying the triangle inequality again gives
\begin{equation}
\label{eq:decomp_grad}
\|g_i-g_j\|_1 \ge \|\mathbb{E}[g_i]-\mathbb{E}[g_j]\|_1 - \|(g_i-\mathbb{E}[g_i])-(g_j-\mathbb{E}[g_j])\|_1.
\end{equation}
By assumption (A4), $\|\mathbb{E}[g_i]-\mathbb{E}[g_j]\|_1 \ge \gamma$, and by (A3),
\[
\|(g_i-\mathbb{E}[g_i])-(g_j-\mathbb{E}[g_j])\|_1 \le \frac{c_1}{\sqrt{n_i}} + \frac{c_1}{\sqrt{n_j}}.
\]
Thus,
\begin{equation}
\label{eq:grad_bound_final}
\|g_i-g_j\|_1 \ge \gamma - \left(\frac{c_1}{\sqrt{n_i}} + \frac{c_1}{\sqrt{n_j}}\right).
\end{equation}
Moreover, from (A3) the noise term satisfies
\[
\|\xi_i-\xi_j\|_1 \le c_2\,\tilde{\sigma}\,\mathcal{C}\,\sqrt{d}.
\]
Define the total clipping deviation as
\[
\mathrm{(clip\_dev)} = \|\epsilon_{\mathrm{clip}}^{(i)}\|_1 + \|\epsilon_{\mathrm{clip}}^{(j)}\|_1.
\]
Substituting these bounds into \eqref{eq:triangle_ineq} yields
\begin{equation}
\label{eq:combined_bound_final}
\|\widetilde{g}_i-\widetilde{g}_j\|_1 \ge \gamma - \left(\frac{c_1}{\sqrt{n_i}} + \frac{c_1}{\sqrt{n_j}}\right) - c_2\,\tilde{\sigma}\,\mathcal{C}\,\sqrt{d} - \mathrm{(clip\_dev)}.
\end{equation}

\textbf{Step 3. Final Separation in Weight Space.} \\[1mm]
Multiplying the bound in \eqref{eq:combined_bound_final} by the learning rate $\eta$ (cf. \eqref{eq:weight_diff_final}), we obtain
\begin{equation}
\label{eq:final_bound_rewrite}
\|w_i-w_j\|_1 \ge \eta \left[\gamma - \left(\frac{c_1}{\sqrt{n_i}} + \frac{c_1}{\sqrt{n_j}} + c_2\,\tilde{\sigma}\,\mathcal{C}\,\sqrt{d} + \mathrm{(clip\_dev)}\right)\right].
\end{equation}
Thus, if
\[
\gamma > \frac{c_1}{\sqrt{n_i}} + \frac{c_1}{\sqrt{n_j}} + c_2\,\tilde{\sigma}\,\mathcal{C}\,\sqrt{d} + \mathrm{(clip\_dev)},
\]
then with high probability,
\[
\|w_i-w_j\|_1 \ge \eta \left[\gamma - \left(\frac{c_1}{\sqrt{n_i}} + \frac{c_1}{\sqrt{n_j}} + c_2\,\tilde{\sigma}\,\mathcal{C}\,\sqrt{d} + \mathrm{(clip\_dev)}\right)\right] > 0.
\]

\paragraph{Discussion.} This lower bound demonstrates that the separation between the updated weights is primarily determined by the intrinsic difference $\gamma$ in the expected gradients of clients with distinct data distributions. Notably, under typical assumptions, $\gamma$ scales linearly with the model dimension $d$, so that as $d$ increases, the separation grows proportionally. Moreover, as the number of training samples $n_i$ increases, the sampling error terms $\frac{c_1}{\sqrt{n_i}}$ and $\frac{c_1}{\sqrt{n_j}}$ shrink, ensuring that $\gamma$ dominates the bound. Consequently, even after incorporating DP noise (with $\sigma = \tilde{\sigma}\,\mathcal{C}$) and clipping errors, the expected gradient difference remains the key factor determining the grouping metric. This validates the use of the $\ell_1$ distance between weight updates as a reliable proxy for distinguishing underlying data distributions, thereby supporting effective clustering.

\subsection{Conclusion}

Under assumptions (A1)–(A4) and with the DP noise scale set as $\sigma = \tilde{\sigma}\,\mathcal{C}$, we have rigorously shown that the $\ell_1$ distance between clients' weight updates satisfies
\[
\|w_i-w_j\|_1 \ge \eta \left[\gamma - \left(\frac{c_1}{\sqrt{n_i}} + \frac{c_1}{\sqrt{n_j}} + c_2\,\tilde{\sigma}\,\mathcal{C}\,\sqrt{d} + \mathrm{(clip\_dev)}\right)\right],
\]
with high probability. In other words, when the intrinsic difference in expected gradients, $\gamma$, exceeds the aggregate sampling, noise, and clipping errors, the clients' weight updates remain well separated. This finding underpins the clustering strategy by confirming that the $\ell_1$ distance between weight updates is an effective proxy for distinguishing the underlying data distributions of clients.

\end{document}